\documentclass[11pt,a4paper]{article}
\usepackage[utf8x]{inputenc}
\usepackage[hebrew,russian,main=english]{babel}
\usepackage[LHE,OT2,LCY,T1]{fontenc}
\usepackage{stfloats}
\usepackage{placeins}
\usepackage[hyperref]{emnlp2020}
\usepackage{cjhebrew}
\usepackage{microtype}
\usepackage{times}
\usepackage{latexsym}
\usepackage{amsmath}
\usepackage{graphicx}
\usepackage{url}
\usepackage{subfig}
\usepackage{adjustbox}
\usepackage{xcolor}
\usepackage{amssymb}
\usepackage{mathptmx}
\usepackage{expex}
\usepackage[title]{appendix}
\usepackage{amsmath}
\usepackage{afterpage}
\usepackage{multirow}
\usepackage{multicol}
\usepackage{booktabs}
\usepackage{epigraph}
\usepackage{enumitem}
\setlist[itemize]{noitemsep, label={\large\textbullet}}
\aclfinalcopy 
\def\aclpaperid{21} 

\title{On the Frailty of Universal POS Tags 
for Neural UD Parsers\\}

\author{Mark Anderson\qquad Carlos G\'{o}mez-Rodr\'{i}guez\\
  Universidade da Coru\~na, CITIC \\
  FASTPARSE Lab, LyS Research Group, \\ 
  Departamento de Ciencias de la Computaci\'{o}n y Tecnolog\'{i}as de la Informaci\'{o}n \\
  Campus Elvi\~{n}a, s/n, 15071 
  A Coru\~{n}a, Spain\\
  \{\texttt{m.anderson,carlos.gomez}\}\texttt{@udc.es}}

\date{}

\newcommand{\mda}[1]{\textcolor{black}{#1}}
\newcommand{\carlos}[1]{\textcolor{black}{#1}}
\newcommand{\carlostwo}[1]{\textcolor{black}{#1}}
\begin{document}
\maketitle
\begin{abstract}
We present an analysis 
on the effect UPOS accuracy has on parsing performance. Results suggest that leveraging UPOS tags as features for neural parsers requires a prohibitively high tagging accuracy and that the use of gold tags offers a non-linear increase in performance, suggesting some sort of exceptionality. We also investigate what aspects of predicted UPOS tags impact parsing accuracy the most, highlighting some potentially meaningful linguistic facets of the problem. 
\end{abstract}

\section{Introduction}
Part-of-speech (POS) tags and dependency parsing have formed a long-standing union in NLP. But equally long-standing has been the question of its efficacy. 
\carlos{Prior to the prevalence of deep learning in NLP, they were shown to be useful for syntactic disambiguation in certain contexts}
\cite{voutilainen1998does,dalrymple2006much,alfared2012pos}. However, for neural network implementations, especially those which utilise character embeddings, POS tags have been shown to be much less useful \cite{ballesteros2015improved,de2017raw}. 

Others have found that POS tags can still have a positive impact when using character representations given that the accuracy of the predicted POS tags used is sufficiently high \cite{dozat2017stanford}. \citet{smith2018investigation} undertook a systematic study of the impact of features for Universal Dependency (UD) parsing and found that using universal POS (UPOS) tags does still offer a marginal improvement for their transition-based neural parser. The use of fine-grained POS tags still seems to garner noticeable improvements 
\cite{ammar2016many}. 

Latterly, POS tags have been commonly utilised implicitly for neural network parsers in multi-learning frameworks where they can be leveraged without the cost of error-propagation \cite{zhang2016stack,yang2017joint,li2018joint,nguyen2018improved,zhang2020pos}. Beyond multi-learning systems, \citet{strzyz2019viable} introduced dependency parsing as sequence labelling by encoding dependencies using relative positions of UPOS tags, thus explicitly requiring them at runtime. 

We follow the work of \citet{smith2018investigation} and evaluate the interplay of word embeddings, character embeddings, and POS tags as features for two modern parsers, one a graph-based parser, Biaffine, and the other a transition-based parser, UUParser \cite{dozat20161,smith201882}. Similar to \citet{zhang2020pos}, we focus on the contribution of POS tags but evaluate UPOS tags.

\paragraph{Contribution} We analyse the effect UPOS accuracy has on two dependency parser systems for a number of UD treebanks. Our results suggest that in order to leverage UPOS tags as explicit features for these neural parsers, a prohibitively high tagging accuracy is needed, and that gold tag annotation seems to possess some exceptionality. We also investigate what aspects of predicted UPOS tags have the most impact on parsing accuracy.

\section{Experimental details}
\carlostwo{We ran three experiments to measure the impact POS\footnote{From this point on we refer to universal POS tags as POS tags rather than UPOS tags for sake of efficiency.} tagging accuracy has on parsing performance when using POS tags as features. Experiment 1 considered the POS tagging accuracy as a controlled variable, set by training taggers as described below and then using the output of these taggers as features for parsers. Experiment 2 was similar, except the size of character embeddings were also changed. Experiment 3 was an extension to test the impact of taggers in an optimal setting where they achieve very high accuracies.}
\paragraph{Data} We use the same subset of UD v2.4 treebanks \cite{nivre2019} as \citet{anderson2020distilling}: Ancient Greek Perseus, Chinese GSD, English EWT, Finnish TDT, Hebrew HTB, Russian GSD, Tamil TTB, Uyghur UDT, and Wolof WTB. We used fastText word embeddings for each language except for Ancient Greek and Wolof \cite{grave2018learning}. For Ancient Greek we use embeddings from \citet{ginter2017conll} and for Wolof those from \citet{heinzerling2018bpemb}. When necessary, we reduced the dimensions 
to 100 using the algorithm of \citet{raunak2017simple}. 

\subsection{Methodology}\label{sec:meth}
\paragraph{POS taggers} We train POS taggers for each treebank separately using the sequence-labelling framework NCRF++ \cite{yang2018ncrf}. We train taggers so as to have POS taggers with varying accuracies ranging from 60 to the maximum score the network can achieve (that fits our binning procedure). The accuracy bins we used were increments of 2.5$\pm$0.3 from 60 to 80 and increments of 1$\pm$0.3 from 80 onwards. We allowed a small window around the desired accuracy for each bin to account for the fact we might never see a model with that exact accuracy. To obtain taggers with varying accuracies, we train each tagger as normal and save models when they reach a certain accuracy. We chose to vary the accuracy of the taggers in this more \textit{natural} way so as to better represent how the taggers would likely behave if they were trained normally but never exceeded the accuracy of a given bin, so it is more likely that easier patterns are learnt first and systematic failures are more likely than if we randomly added noise.

\paragraph{Network details} We use the default parameters for both parsers, i.e. those reported in each subsequent paper. We use v2.3 of UUParser\footnote{\href{https://github.com/UppsalaNLP/uuparser/blob/master/README.md}{UUParser GitHub} from \citet{smith201882}.} and use a PyTorch implementation of Biaffine.\footnote{\href{https://github.com/yzhangcs/parser}{Biaffine PyTorch GitHub} based on \citet{dozat2017stanford}.} The features to the networks are the word embeddings as mentioned above, character embeddings, and POS tag embeddings, with the latter two embeddings being randomly initialised. For Experiment 1, the character embedding size was 32 and varied as specified below for Experiment 2. The BiLSTM output dimension of the character embedding layer was 100 and the embedding dimension of the word and POS embeddings were also 100. These dimensions were chosen to control the contribution from each feature, but it is obviously feasible that optimising these contributions could result in different absolute results. However, keeping these static unless purposefully changing them for controlled input means we can make relative comparisons.

\paragraph{Experiment 1} We trained parsers for each treebank with gold tags and with predicted tags using a subset of the POS taggers with accuracy bins 60, 70, 80, 86, 91, and 93. The values were chosen such that we could cover a reasonable range and include as many treebanks as possible (e.g. only English, Hebrew, and Russian have taggers which achieve 93\% accuracy). \carlos{The parsers trained with predicted tags are run on inputs tagged by the same model, and those trained with gold tags are tested both on gold and a range of predicted tags.} The goal of this experiment was to test the sensitivity of parsers to POS tagging accuracy for different treebanks. We also trained parsers without POS tags as a baseline for comparison.

\paragraph{Experiment 2} We trained parsers for each treebank with gold tags and with predicted tags using a subset of the POS taggers with accuracy bins 80, 86, and the max accuracy for each treebank which was on average 91(3). \carlos{Each parser is run on inputs tagged by the same model. We used}
varying character embedding sizes of 32, 100, 180, 325, and 500. We also train parsers with these varying character embedding sizes with no POS tags as a baseline.

\paragraph{Experiment 3} We trained parsers with and without predicted POS tags for treebanks for which we obtained high-scoring POS taggers with a mean accuracy of 96(2) to evaluate the trend observed in Experiment 1. We use the settings from Experiment 1. The treebanks used were Catalan AnCora, Japanese GSD, Latin ITTB, and Polish PDB.

\section{Results and analysis}



\begin{figure*}[t!]
    \centering
    \includegraphics[width=0.99\linewidth]{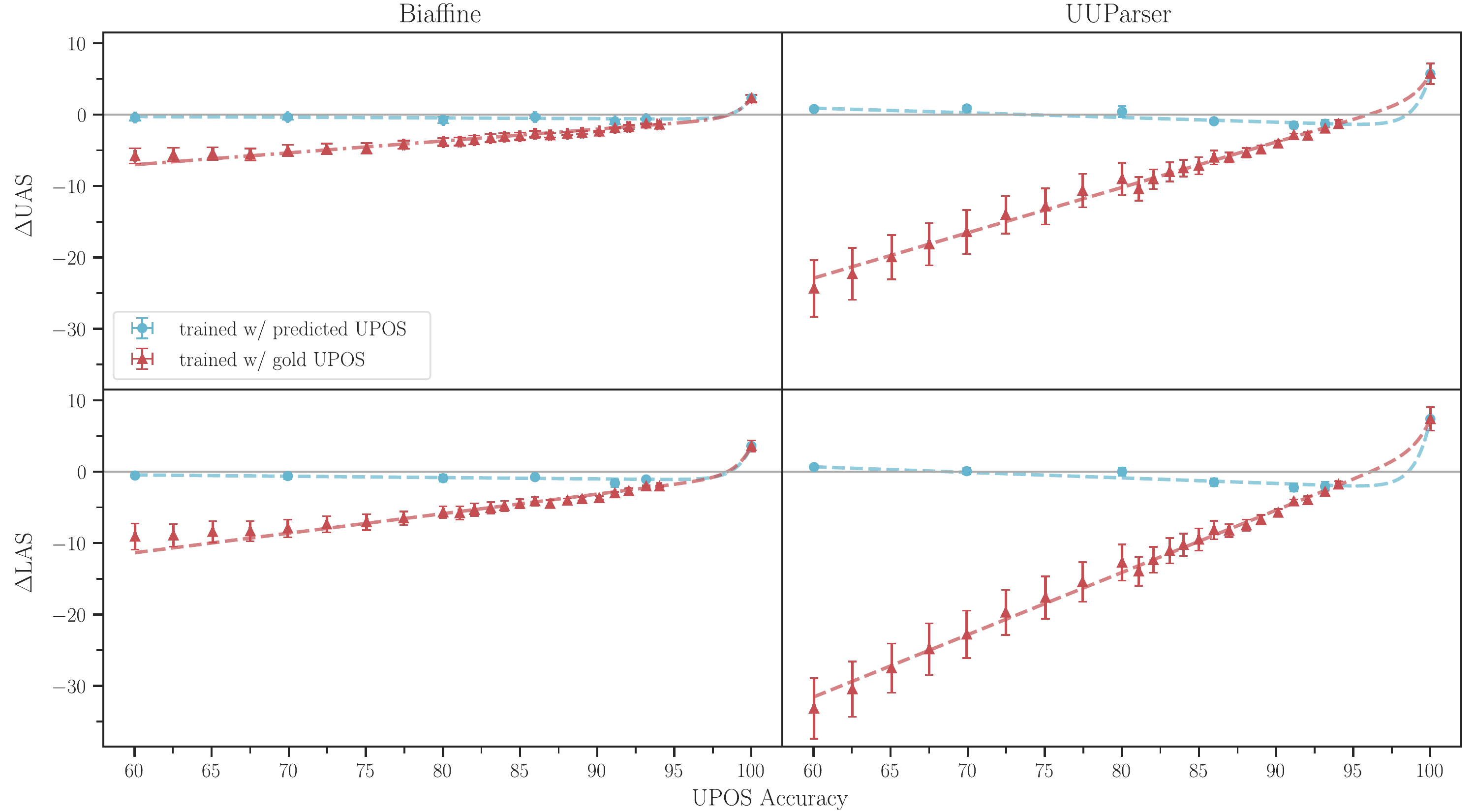}
    \caption{Average $\Delta$ attachment scores across all treebanks over the relative baseline parsers trained without POS tags, plotted with respect to POS tag accuracy for parsers trained with predicted (blue, circles) and gold (red, triangles) tags.}
    \label{fig:delta_accs}
\end{figure*}
\paragraph{Experiment 1} The results of Experiment 1 are shown in Figure \ref{fig:delta_accs}, where the average difference in attachment scores between the baseline parsers (without POS tags) and those with differing POS tag accuracies are shown. We show the differences in attachment scores rather than the absolute values, as averaging over treebanks obscures differences. 

There is an unsurprising relation between parsing score and tagging performance when training with gold POS tags. What is less expected is how little of an impact is observed when using predicted tags during training, with an almost consistent performance with respect to POS tag accuracy.

The gold training trend for the graph-based parser suggests that it is less sensitive to POS tag accuracy than the transition-based parser. This is likely due to the transition-based parser being able to leverage POS tags more, so that it will see more of an impact when tagging accuracy is low. This is somewhat corroborated by the larger positive difference over the baseline when using gold tags at prediction time for UUParser compared to the increase seen for Biaffine.

Another notable phenomenon is that the results for parsing texts annotated with gold POS (rightmost point in each plot) outperform what could be expected from extrapolating the general trends. 
This raises the question as to whether this is due to smooth nonlinear accuracy increases in the rightmost part of the curves (where we couldn't obtain taggers) 
 or to a sudden jump at the very end in a hockey-stick shape, indicating an exceptionality of gold POS tags and inadequacy of even very accurate but imperfect POS tags (which is relevant under the assumption that tagging accuracy can be pushed further with future model and/or training data improvements). Answering this question was the motivation for Experiment 3.

Almost exclusively, using predicted POS tags does not outperform the parser trained without any POS tags. Curiously, the only parsers that are marginally better 
are those trained with predicted POS tags from the least accurate POS taggers.  

Figures \ref{fig:uas_biaffine}\nobreakdash--\ref{fig:las_uppsala} in the Appendix show these results for each treebank separately, and almost all treebanks follow the general trend seen for both parsers. The only exception is Tamil TTB for UUParser, which benefits from POS tags both when training with gold and predicted tags. 
\carlos{Tamil TTB is the smallest treebank, and it has the additional difficulty for parsing and tagging of being an agglutinative language, so possibly this combination of factors lends itself well to leveraging POS tags even in less than optimal circumstances.}
Tamil is also the lowest performing language with respect to POS tagging and parsing accuracy, but compared to Uyghur and Ancient Greek (the next two lowest performing languages) it outperforms both when using gold tags. In fact, Tamil has the biggest difference when using gold tags over the baseline than any other language, suggesting that they might be particularly useful when there is a heightened probability of ambiguity coupled with a dearth of data. 

\begin{figure*}[t!]
    \centering
    \includegraphics[width=0.99\linewidth]{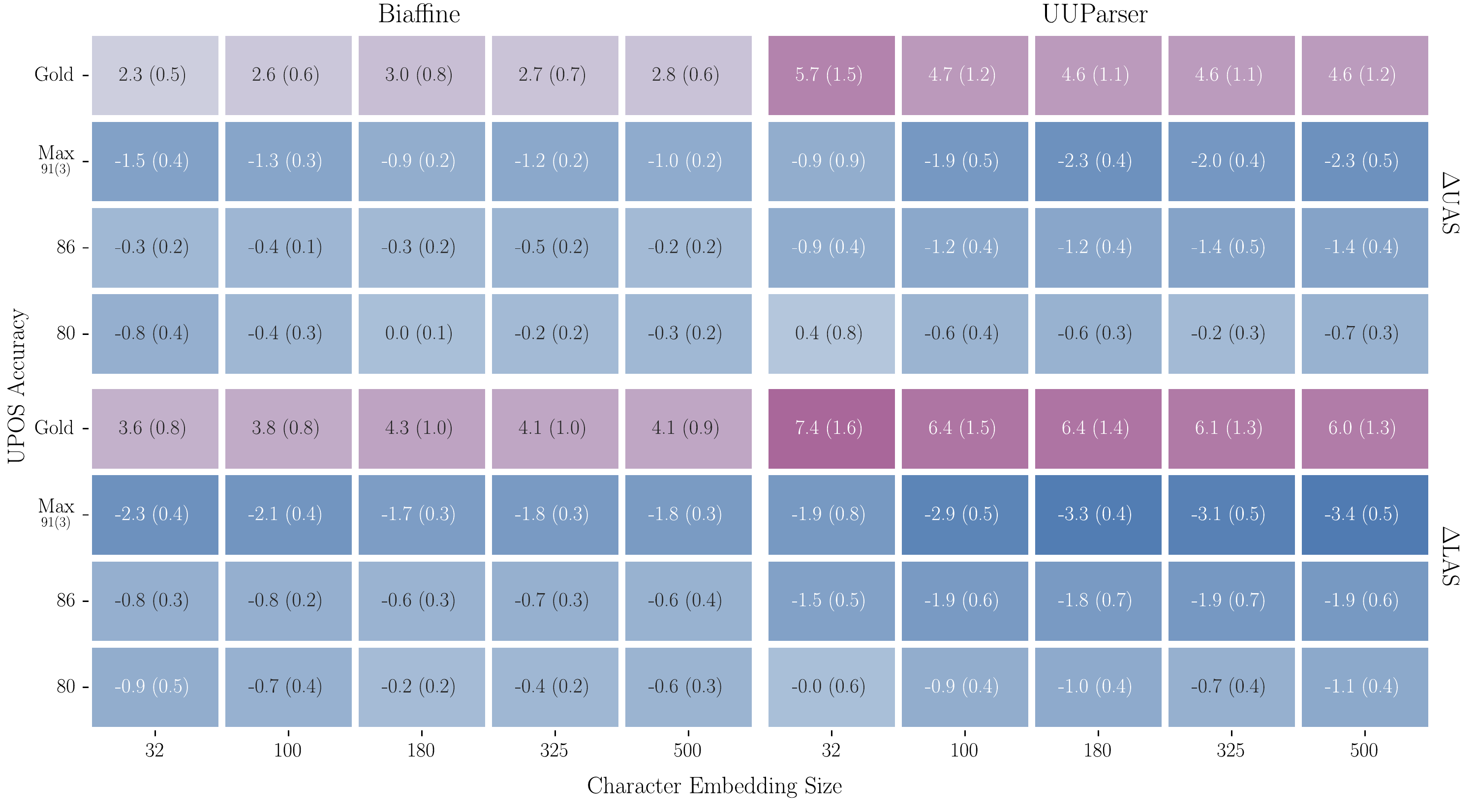}
    \caption{Average $\Delta$ attachment scores across all treebanks over the relative baseline parsers trained without POS tags for different character embedding sizes and different POS tag accuracies (80\%, 86\%, max (average of 91\%) tagger accuracy for each treebank, and gold).}
    \label{fig:delta_accs_chars}
\end{figure*}

\begin{figure}[tb]
    \centering
    \includegraphics[width=0.98\linewidth]{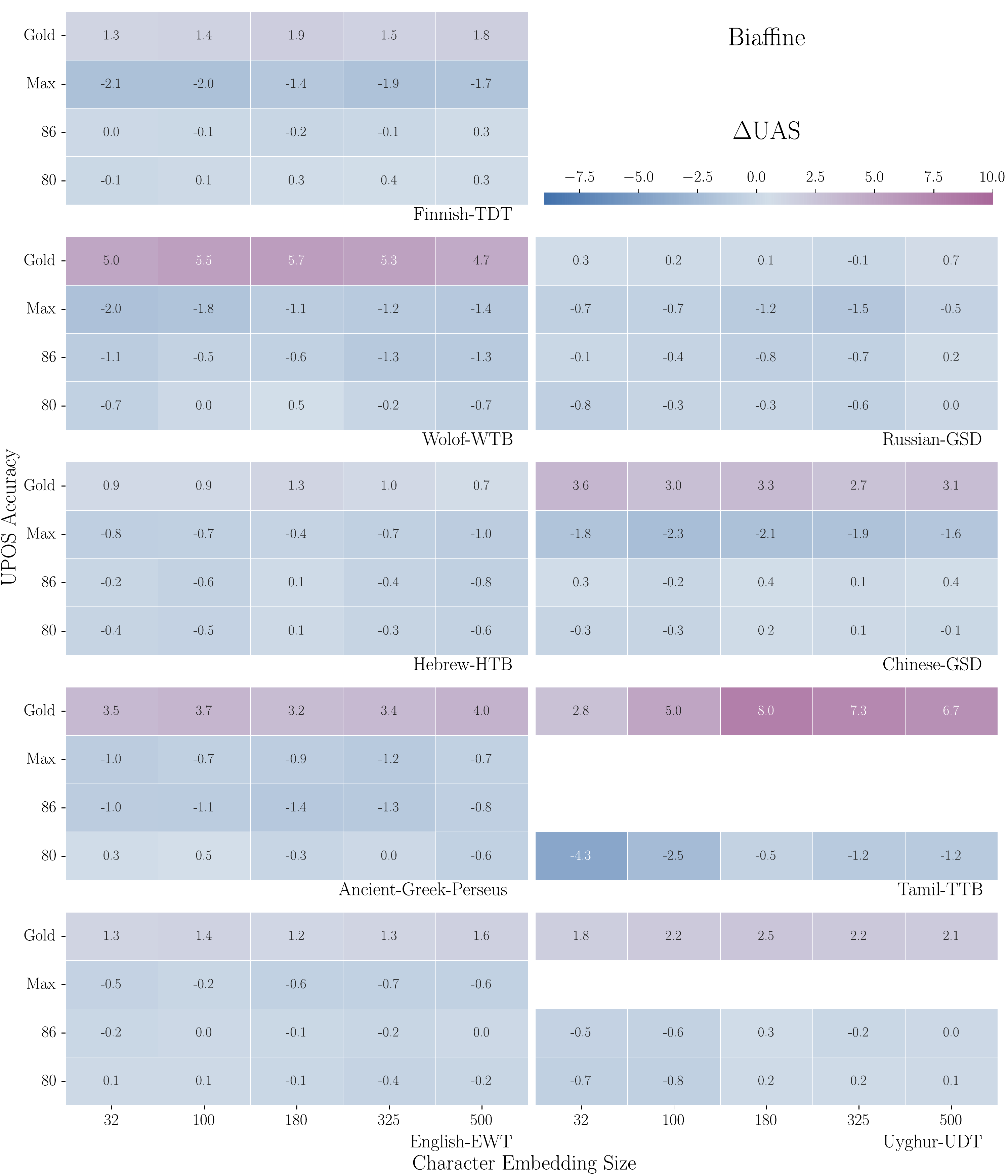}
    \caption{$\Delta$ UAS for each treebank for Biaffine compared to the baseline parsers trained without POS tags for different character embedding sizes and different POS tag accuracies.}
    \label{fig:uas_biaffine_heat}
\end{figure}

\begin{figure}[tb]
    \centering
    \includegraphics[width=0.98\linewidth]{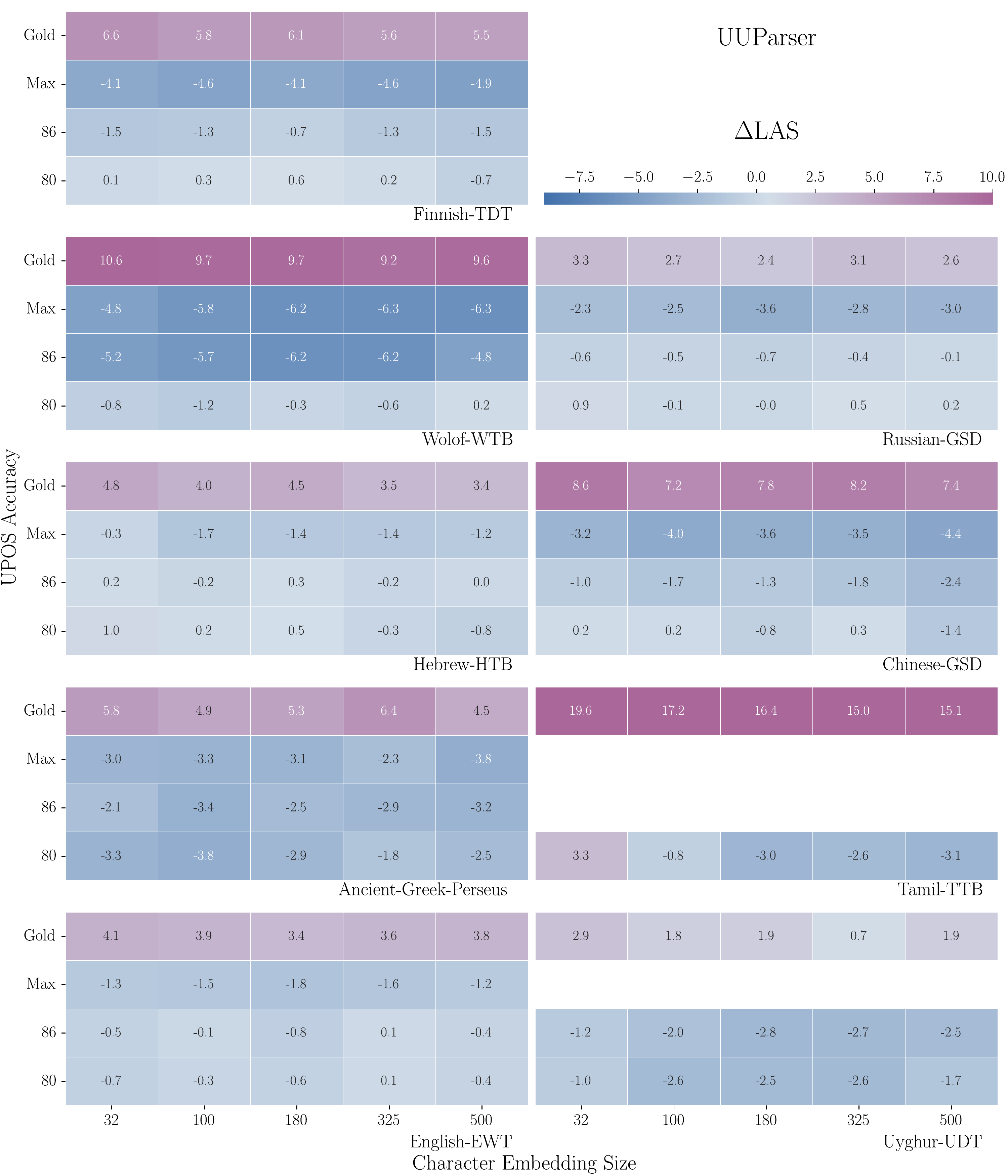}
    \caption{$\Delta$ LAS for each treebank for UUParser compared to the baseline parsers trained without POS tags for different character embedding sizes and different POS tag accuracies.}
    \label{fig:las_uppsala_heat}
\end{figure}

\paragraph{Experiment 2} The average attachment score differences for Experiment 2 are shown in Figure \ref{fig:delta_accs_chars}. 
 This experiment was initially devised as we anticipated POS tags would have more of a positive effect, especially for higher accuracy taggers, and we wanted to evaluate if having larger character embeddings would offset this. However, as the results of Experiment 1 showed no improvement over not using POS tags at all, this experiment became a verification of the inutility of predicted POS tags instead. And it is clear that in all contexts where predicted tags are used, no matter what the character embedding size is or what parser is used, predicted POS tags perform worse than not using POS tags at all. 
\begin{table}[b!]
    \centering
    \tabcolsep=.1cm
    \small
    \begin{tabular}{lccccc}
    \toprule
     & \multicolumn{2}{c}{Biaffine}& \phantom{} & \multicolumn{2}{c}{UUParser} \\\cmidrule{2-3}\cmidrule{5-6}
    \textbf{Char} & \multicolumn{1}{c}{UAS} & LAS & \phantom{} & UAS  & LAS\\
   \textbf{32}&84.0 (5.9)&78.6 (8.7)&&77.9 (8.3)&71.9 (10.0)\\
\textbf{100}&83.9 (6.3)&78.6 (8.9)&&79.0 (7.3)&73.0 (9.3)\\
\textbf{180}&83.4 (6.8)&78.1 (9.4)&&79.1 (7.2)&73.1 (9.2)\\
\textbf{325}&83.6 (6.7)&78.1 (9.5)&&79.2 (7.1)&73.3 (8.9)\\
\textbf{500}&83.6 (6.4)&78.1 (9.1)&&79.3 (7.0)&73.4 (8.8)\\
     \bottomrule
    \end{tabular}
    \caption{Average attachment scores for different character embedding sizes (Char) 
    without POS tags.}
    \label{tab:char_better}
\end{table}
The unexpected dip in performance as tagging accuracy increases is even clearer here, as this trend is consistent across different character embedding sizes and is the case for both parsers. This decrease in performance is even more marked as the performance actually increases when increasing the character embedding size and not using POS tags at all for UUParser, as shown in Table \ref{tab:char_better}. This result corroborates one of the many observations from \citet{smith2018investigation}. For the graph-based parser there is a negligible negative impact at higher character embedding sizes. Both parser implementations use a BiLSTM to create the character vector input to the network, so this seems more likely to be a result of the transition-based decoder leveraging features more than the graph-based one. The transition-based parser's ability to leverage POS tags in optimal settings is even clearer in Figure \ref{fig:delta_accs_chars}, as UUParser has twice the improvement using gold tags than that of Biaffine. Also, the impact of predicted POS tags is more pronounced as character embedding sizes increase for UUParser, but 
for Biaffine there is only 
a slight tendency to decrease as the character embedding size increases. 
\begin{figure*}[b!]
    \centering
    \includegraphics[width=0.99\linewidth]{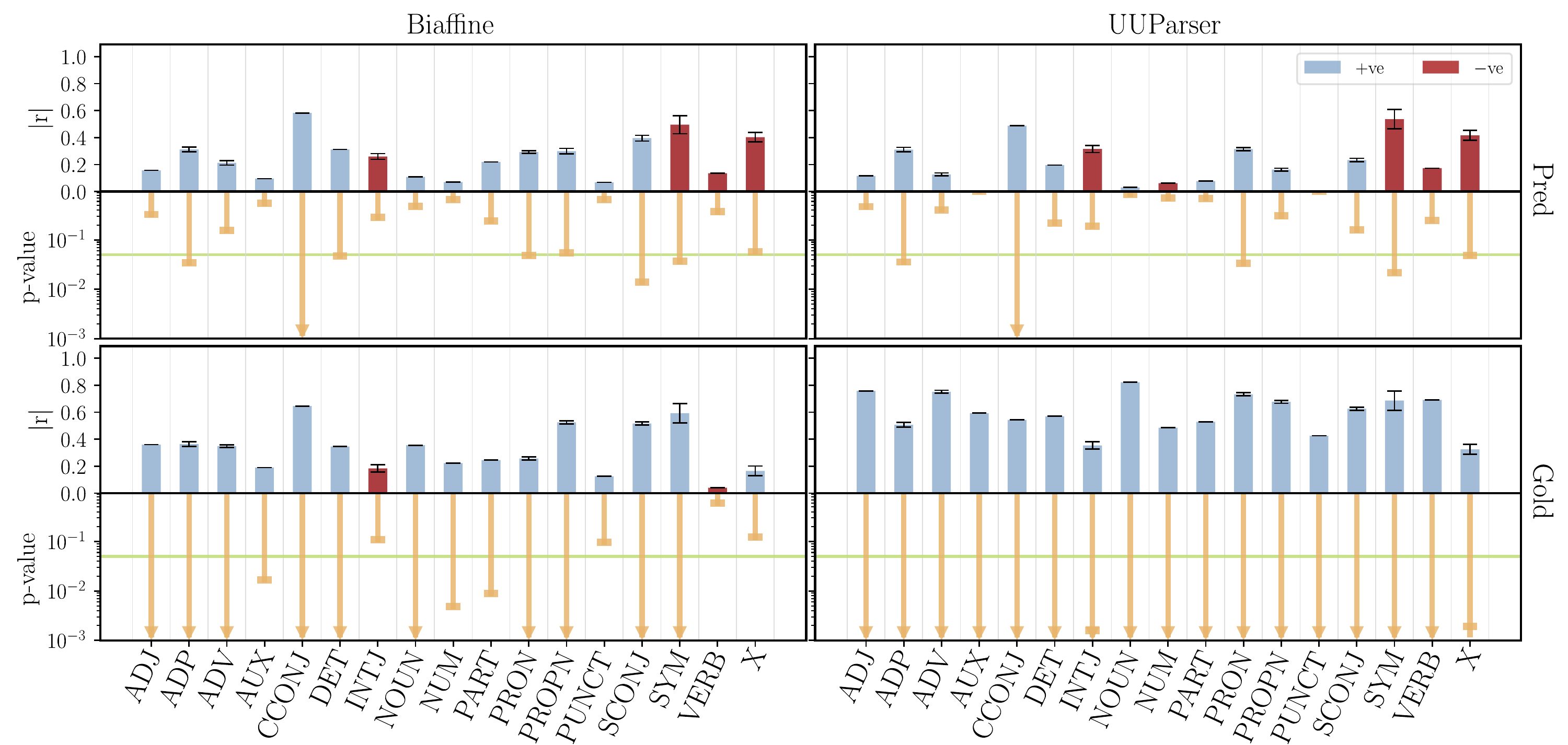}
    \caption{Pearson coefficients for the F1-score for separate POS tags and global LAS where positive (+ve) coefficients are shown in blue and negative (-ve) are shown in red. The corresponding p-values are shown below (orange) where an arrow head means the value was below 0.001. Left subplots are for Biaffine parsers, right for UUParsers, top row is for parsers trained with predicted tags, and bottom for parsers trained with gold tags.}
    \label{fig:corr_f1}
\end{figure*}
We show the breakdown for each treebank in Figures \ref{fig:uas_biaffine_heat}\nobreakdash--\ref{fig:las_uppsala_heat} where again Tamil is clearly an outlier for UUParser, as it is the only language where any settings with predicted POS tags result in a positive increase (80 POS tag accuracy, character embedding size of 32) and has by far and away the largest increase when using gold tags (a factor of 2 greater than the next best improving language, Wolof, for both UAS and LAS).
\paragraph{Experiment 3} In Figure \ref{fig:delta_accs}, there is a point around 
96-98 POS tag accuracy where the parsers outperform the baselines  without POS tags. Due to a lack of models in that range, this is just an extrapolation. So we trained parsers 
with treebanks, listed above 
in the description of Experiment 3, for which we could obtain high POS tagging accuracy. The results of these parsers are shown in Table \ref{tab:limit_table}. Only the top two treebanks with the highest tagging accuracy (Catalan AnCora and Japanese GSD) perform better than using no POS tags, and only for UUParser. However, when the performance is below the baseline the difference is marginal.
\carlos{These results are consistent with the 
extrapolations in Figure \ref{fig:delta_accs} and suggest a sharp increase in the $\Delta$ attachment score slopes when POS tagging accuracy is in the 98-100 range, i.e., that predicted POS tags suddenly start being useful when they are very close to gold POS tags. This suggests that there may be certain tag patterns or contexts that are particularly relevant for parsing, but especially difficult for taggers to learn.}

\begin{table}[t!]
    \centering
    \tabcolsep=.1cm
    \small
    \begin{tabular}{rccccccc}
    \toprule
     & \multicolumn{2}{c}{Biaffine}& \phantom{} & \multicolumn{2}{c}{UUParser} & \multicolumn{1}{c}{}\\\cmidrule{2-3}\cmidrule{5-6}
          & \multicolumn{1}{c}{UAS} & LAS & \phantom{} & UAS  & LAS &  POS$_{ACC}$  \\\midrule
        \multirow{1}{*}{\textit{Catalan-AnCora}} \\
         Predicted & \multicolumn{1}{c}{92.59} & 89.57  & \phantom{} & \textbf{90.88} & \textbf{88.03} & 98.26 \\
         None & \textbf{92.89}  & \textbf{90.33} & \phantom{} & 90.82 & 87.92 & n/a\\
        \multirow{1}{*}{\textit{Japanese-GSD}}\\
         Predicted &\multicolumn{1}{c}{95.02} & 93.66  & \phantom{} & \textbf{94.56} & \textbf{92.94}  & 97.69\\
         None & \textbf{95.12} & \textbf{93.54} & \phantom{} & 94.47 & 92.74 & n/a \\
        \multirow{1}{*}{\textit{Polish-PDB}}\\
         Predicted & \multicolumn{1}{c}{92.78} & 89.97 & \phantom{} &  89.25 & 85.57  & 97.52\\
 None & \textbf{93.64} & \textbf{90.94} & \phantom{} & \textbf{89.32} & \textbf{85.60} & n/a\\ 
 \multirow{1}{*}{\textit{Latin-ITTB}}\\
         Predicted & \multicolumn{1}{c}{90.92}& 88.47  & \phantom{} & 86.99  & 83.99  & 97.46\\
         None & \textbf{91.09} & \textbf{88.74} & \phantom{} & \textbf{87.25} & \textbf{84.25} & n/a \\
        \bottomrule
    \end{tabular}
    \caption{Performance for treebanks with high scoring POS taggers 
    trained with predicted POS tags\carlostwo{, compared to the performance on the same treebanks without using POS tags}.}
    \label{tab:limit_table}
\end{table}
\begin{figure*}[t!]
    \centering
    \includegraphics[width=0.99\linewidth]{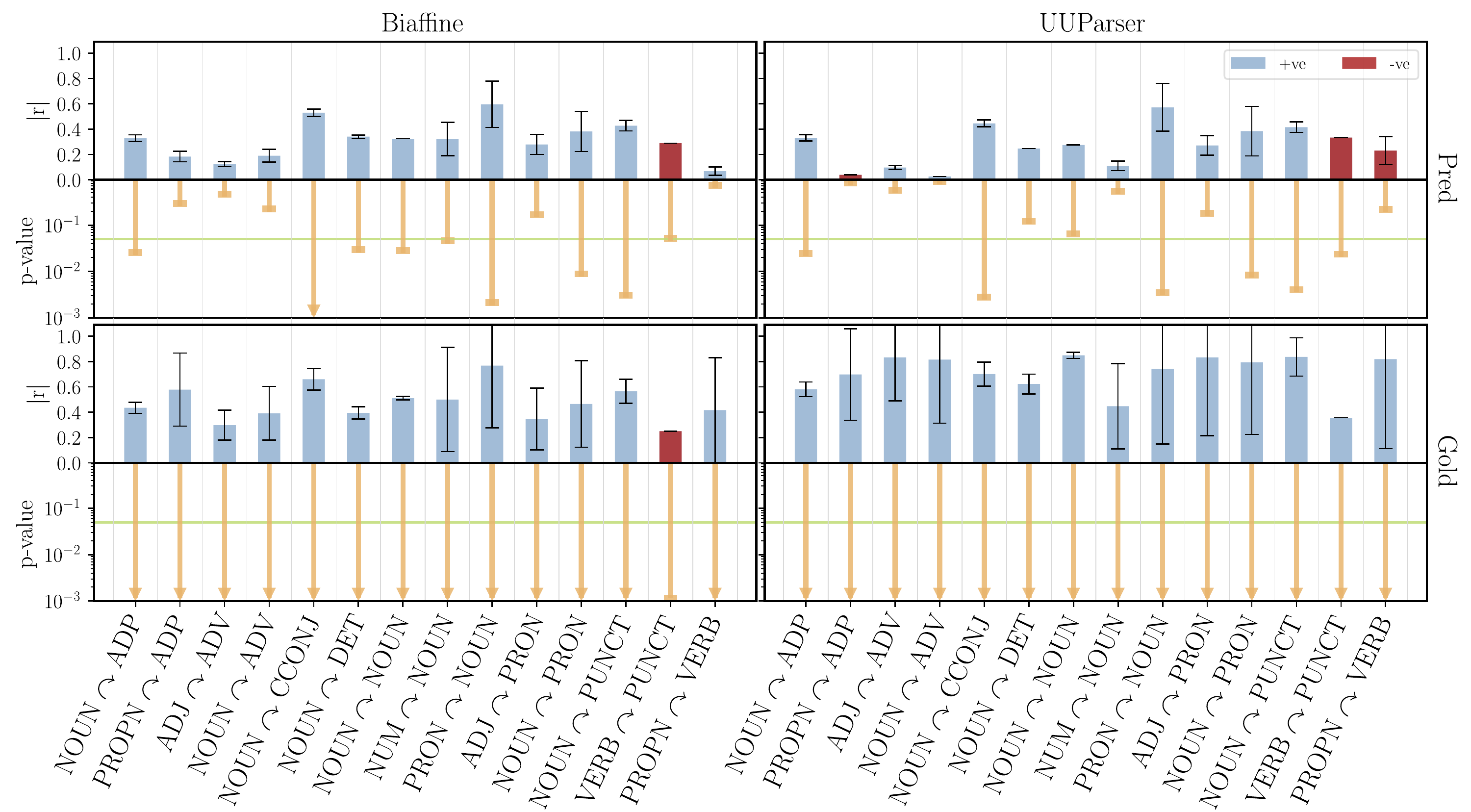}
    \caption{Pearson coefficients for the tagging F1-score for child$\curvearrowright$head pairs and global LAS where positive (+ve) coefficients are shown in blue and negative (-ve) are shown in red. The corresponding p-values are shown below (orange) where an arrow head means the value was below 0.001. Left subplots are for Biaffine parsers, right for UUParsers, top row is for parsers trained with predicted tags, and bottom for parsers trained with gold tags.}
    \label{fig:corr_pair}
\end{figure*}

\subsection{Parsing difficulty of POS tags}
We then delved deeper by looking at the difficulty of predicting arcs and labels for each POS tag type. The full results are shown in Figures \ref{fig:duas_bar} and \ref{fig:dlas_bar} in the Appendix, where the average differences in score with respect to the baseline model (no POS tags) are given. \texttt{X} (the UPOS tag for ``other'') is consistently difficult across parsers and parser types, except that the loss in UAS for UUParser is much smaller than for Biaffine for both training with predicted and gold tags. However, the only time \texttt{X} is consistently better than the baseline model for LAS is when using gold tags at runtime and only with UUParser.

Another noticeable feature in these results is that for the max POS predicted accuracy and gold tag parsers for UUParser, \texttt{INTJ} (interjection) performs significantly better, both using predicted POS tags and gold tags for training, compared to the lower POS tag accuracy parsers. 
Beyond this, the performances 
echo the global scores with respect to the tagging accuracy. 

Next, we evaluated the correlations (Pearson coefficient) between tagging accuracy for each POS tag and global parsing performance. For these correlation results and all those that follow,  
 we use the same taggers and parsers from Experiment 1.
We only report results for LAS 
for the sake of space.

Figure \ref{fig:corr_f1} shows the Pearson coefficient with the corresponding p-value for the correlations between the F1-score for each POS tag and the global LAS score for both Biaffine and UUParser, for both predicted and gold POS tags used in training. Training with gold tags, the accuracy for every tag is positively correlated with parsing performance for UUParser and the correlations are all statistically meaningful. The correlations range from about 0.4 (\texttt{INTJ} and \texttt{X}) to about 0.8 (\texttt{ADJ}, \texttt{ADV}, \texttt{NOUN}, and \texttt{PRON}). For Biaffine, the correlations are much weaker ranging from 0.2 (\texttt{AUX}) to 0.6 (\texttt{CCONJ}, coordinating conjunction, and \texttt{SYM}, symbol) for those which are statistically significant.

For the systems trained with predicted POS tags, the correlations are much weaker for UUParser and only 5 are statistically significant. UUParser and Biaffine have much more similar correlations under these settings, where Biaffine has 2 other tags significantly correlated but its set contains those of UPParser. Of those that are significantly correlated for both, \texttt{SYM} and \texttt{X} are actually negatively correlated, suggesting that the taggers either fail to generalise or fail to capture certain tagging patterns. A noticeable exception is the \texttt{CCONJ} tag which is both strongly correlated (about 0.6 for both) and statistically significant for both parsers. This is likely due to the nature of \texttt{conj} relations, where 
\carlos{dependents are}
connected to the conjunct rather than the head of the conjunct 
\carlos{(e.g. the second conjoined object of a verb is connected to the first)}
and so 
\carlos{should be parsed differently}
than if they occurred without a \texttt{CCONJ}. 

\begin{figure*}[t!]
    \centering
    \includegraphics[width=0.99\linewidth]{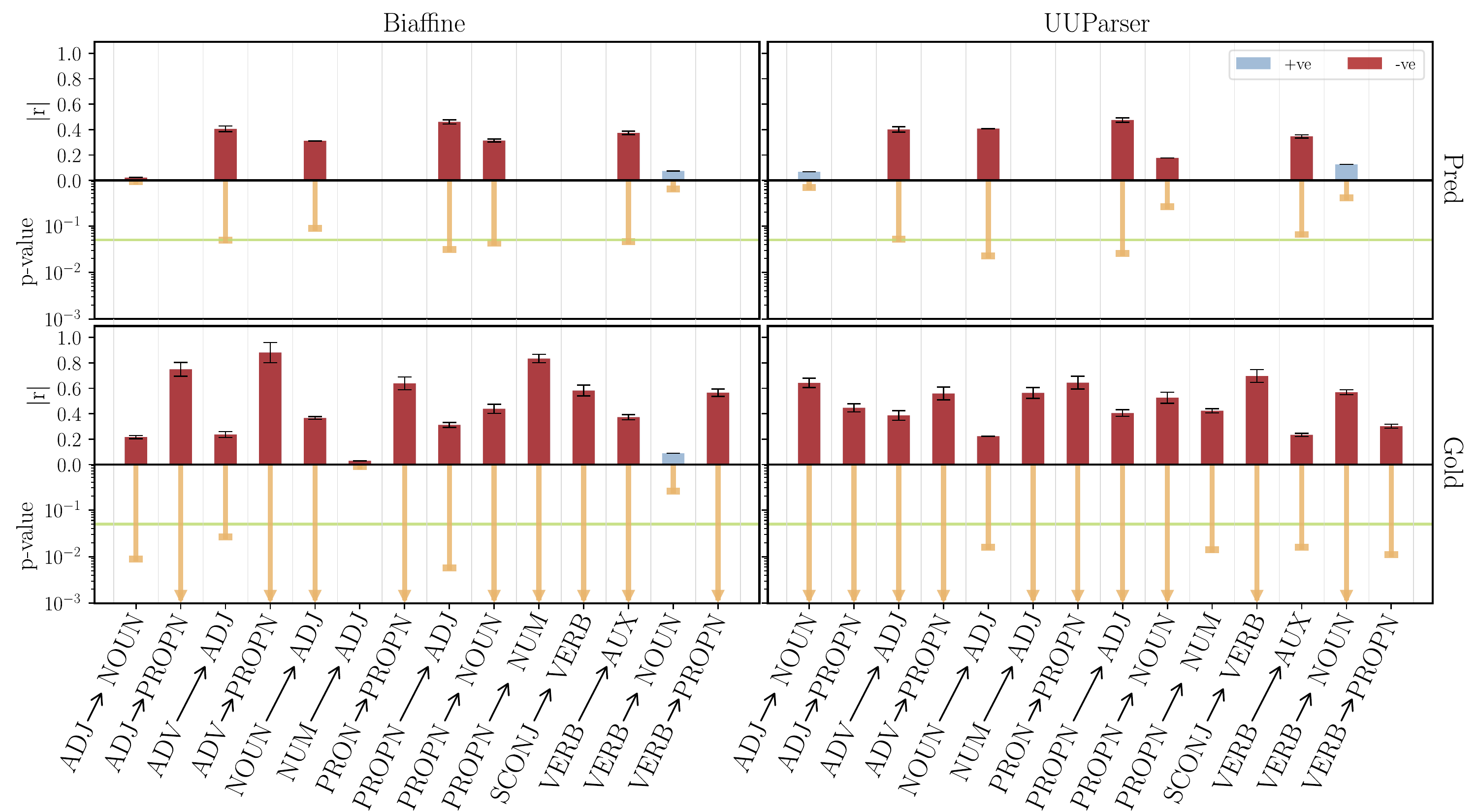}
    \caption{Pearson coefficients for the error rate of individual error types \texttt{POS}$_\textrm{X}\rightarrow$\texttt{POS}$_\textrm{Y}$ and global LAS where positive (+ve) coefficients are shown in blue and negative (-ve) are shown in red. The corresponding p-values are shown below (orange) where an arrow head means the value was below 0.001.  Left subplots are for Biaffine parsers, right for UUParsers, top row is for parsers trained with predicted tags, and bottom for parsers trained with gold tags.}
    \label{fig:corr_error}
\end{figure*}

Figure \ref{fig:corr_head} in the Appendix 
shows the correlation for the tagging accuracy of the head for each tag type. Across all systems, there is a correlation for the head of \texttt{INTJ} nodes 
(0.7 for predicted training, 0.5 for gold). This is perhaps due to \texttt{INTJ} nodes typically being attached to \texttt{VERB} or \texttt{NOUN} nodes, and that this narrow context means that the parsers will always look for a node like these and if the correct node is incorrectly tagged, this could disrupt the arc predictions and would be better off without the tagging information. 

\texttt{ADP} (adposition) nodes are similar but with a lower correlation (about 0.4 for all systems). And again this might be due to these nodes occuring in less diverse contexts. 
\texttt{X} nodes are strongly negatively correlated for Biaffine for both gold and predicted training systems (0.6 and and 0.8 respectively) and similarly \texttt{SCONJ} (subordinating conjunction) nodes (0.7 and 0.5). 
Perhaps the diversity of the contexts in which these tags occur makes it difficult for the parser \carlos{to leverage POS information}. 
\texttt{ADV} nodes follow a similar trend, being negatively correlated for 3 of the 4 systems (gold UUParser being the exception),
\carlos{which could also be related to diversity of contexts:}
adverbs such as \textit{very} never attach to verbs, but to other adverbs, and often the use of \texttt{ADV} covers 
situations where a word doesn't satisfy the definition of another POS tag.

Figure \ref{fig:corr_pair} shows the correlation of combining the accuracy of POS tags and the tags that govern them with global LAS scores. Only pairs that occur 10 times in 4 treebanks are included. The union of pairs with the highest correlations across all systems are shown (the 10 most highly correlated and statistically significant for each parser). The correlations are positive with one exception of \texttt{PUNCT} nodes headed by \texttt{VERB} nodes, which are weakly negative for all systems except gold UUParser. 
Conversely, \texttt{PUNCT} nodes headed by \texttt{NOUN} nodes have positive correlations for all systems (0.4 for all except gold UUParser which is about 0.8). Other than this, \texttt{CCONJ} nodes headed by \texttt{NOUN} nodes are positively correlated (0.6-0.7) for all systems, which adds to the discussion above regarding \texttt{CCONJ} tags and suggests that it helps more specifically when conjuncts are \texttt{NOUN} nodes.

\subsection{Dependency distance}
Figures \ref{fig:uas_dep_dist_biaffine}\nobreakdash--\ref{fig:uas_dep_dist_uppsala_gold} in the Appendix show the attachment scores and occurence rates for each POS tag in dependency distance bins. 
Most tags decrease in performance as the distance increases. Other than \texttt{NOUN}, \texttt{PUNCT}, and \texttt{VERB}, the occurrence of longer-distanced edges are significantly lower than short-distanced ones. Of these, \texttt{NOUN} has a much more significant drop in performance as distance increases 
across all systems.

Figure \ref{fig:corr_dist} in the Appendix shows 
the combinations of POS tag and dependency distance with highest correlation with LAS.
\texttt{CCONJ} appears in 8 pairs (out of 24) and appears 3 out of 6 times for the distances of 3 or less. This further supports the findings from above that awareness of \texttt{CCONJ} nodes is especially beneficial.  Beyond this, most pairs (19) have distances of 4 or greater which is larger than the mean dependency distance typically observed in natural languages, e.g. it is 3.6 (0.4) averaged over all treebanks in UD v.2.4 using the equation from \citet{liu2008dependency}. 

\subsection{Error types}
Finally, we evaluated which type of tagging errors are the most likely to impact parsing performance. In Figures \ref{fig:uas_x_as_y_biaffine}\nobreakdash--\ref{fig:uas_x_as_y_uppsala_gold} in the Appendix, we show the corresponding attachment scores and counts of each error of tagging a gold tag of \texttt{POS}$_\textrm{X}$ as \texttt{POS}$_\textrm{Y}$ in confusion matrices for both parser types, for training with gold tags and with predicted tags with taggers of accuracy 80, 86, 91(3). We include these 
to supplement the following analysis and allow for comparisons to other error types that aren't shown. 
However, we can see that a lower occurrence rate of errors is associated with lower attachment scores and errors have a larger impact on LAS than UAS.

Figure \ref{fig:corr_error} shows the highest correlated and statistically significant tagging errors. Correlations are between the error rate and the global LAS scores. Only error types that occur 10 times in the output of at least two taggers for at least 4 treebanks are included (the 5 most correlated for each parser). This is due to the fact that looking at the correlation between error rates and LAS when an error type rarely occurs will still give \textit{statistically meaningful} correlations, as the absence of stats is one step removed from the correlation calculation. Error types are negatively correlated with parsing performance (the exceptions are those which aren't statistically significant for some systems). 
Correlations are strongest when training with gold POS tags. 
For Biaffine they are either much stronger or much weaker than UUParser, e.g. \texttt{ADJ}$\rightarrow$\texttt{PROPN} is over 0.8 whereas it is only about 0.5 for UUParser, 
\texttt{PROPN}$\rightarrow$\texttt{NUM} is about 0.8 for Biaffine and about 0.4 for UPParser. In contrast, \texttt{ADJ}$\rightarrow$\texttt{NOUN} and \texttt{ADV}$\rightarrow$\texttt{ADJ} are only about 0.2 for Biaffine but are about 0.6 and 0.4, respectively, for UUParser. 

Two POS tag pairs appear in error types where both directions are observed, \texttt{PROPN}$\leftrightarrow$\texttt{ADJ} and \texttt{NOUN}$\leftrightarrow$\texttt{ADJ} for both parsers trained with gold tags. For the former, it appears that qualifiers that refer to nations or groups are often problematic as a similar form or the same one appear as \texttt{PROPN} and \texttt{ADJ}, e.g. \textit{Sunni}, \textit{African}, \textit{Mexican}. For the error type \texttt{ADJ}$\rightarrow$\texttt{PROPN}, another issue seems to be the capitalisation of certain words which either appear on their own or with limited punctuation, e.g \textit{Wonderful!}, \textit{Marvelous!}, or refer to something fixed but not quite a named entity, e.g. \textit{Parliamentary elections}, \textit{Perfect Score}. This is the case in English, and we apologise for the Anglo-bias, but the author isn't proficient in the other languages used. However, these errors do occur at a general level. 
It appears to be similar in Russian (\foreignlanguage{russian}{Бургундского} \textit{- Burgundy}, \foreignlanguage{russian}{Гомельская} - \textit{Gomel Region}); Finnish (\textit{Suomalaisen - Finnish}, \textit{eurooppalaisen - European}); and in Hebrew (\cjRL{'y.tlqyt} - \textit{Italian}, \cjRL{grmnyt} - \textit{German}). 
The only language where neither of these error types occur is Wolof as it doesn't have an adjective category \cite{dione2019developing}.

For the other bidirectional error type (\texttt{NOUN}$\leftrightarrow$\texttt{ADJ}), there appears a similar issue for \texttt{ADJ}$\rightarrow$\texttt{NOUN} as \texttt{ADJ}$\leftrightarrow$\texttt{PROPN} for nations or groups, but \texttt{NOUN} is used instead of \texttt{PROPN}. Beyond this, when a \texttt{NOUN} is incorrectly tagged as an \texttt{ADJ} this occurs 44.7 (14.4)\% when it is governed by another \texttt{NOUN}. This is especially prominent for English and Hebrew (65.8\% and 64.4\% respectively) with the lowest rate occurring for Ancient Greek and Tamil (25.8\% and 28.9\% respectively). The issue of tagging \texttt{NOUN} tokens governed by another \texttt{NOUN} token is also apparent in Figure \ref{fig:corr_pair} where this pair has a correlation coefficient of about 0.4 for both Biaffine systems and 0.8 for the gold trained UUParser system (for the predicted POS tag UUParser system, it isn't statistically significant). Again Wolof is an outlier as the error \texttt{NOUN}$\rightarrow$\texttt{ADJ} never occurs, presumably because it never has any \texttt{ADJ} tokens to learn. 

Only two error types are statistically significant for all systems: \texttt{ADV}$\rightarrow$\texttt{ADJ} and \texttt{PROPN}$\rightarrow$\texttt{ADJ}, the latter having been discussed above. The former isn't particularly prevalent, occurring with an error rate of 5.8 (5.7)\% on average across all languages (\mda{except Wolof}) with Russian and Tamil having the highest rates (15.5\% and  15.0\%, respectively) and   Chinese and Hebrew having the lowest (0.6\% and 1.7\%). For English at least, two issues are clear. Words that have the same form when used as an adverb or adjective are commonly mis-tagged as \texttt{ADJ} when they should be \texttt{ADV}, e.g. \textit{more}, \textit{worst}, \textit{better}, and so on. And also when an adverb is used in hyphenated adjectival phrases such as \textit{fully} in \textit{fully-fledged} and \textit{ill} in \textit{ill-advised}. 

As Wolof was such an outlier with respect to common tagging errors (with those that impacted parsing performance) we looked at those most common in Wolof. \texttt{DET}$\rightarrow$\texttt{TAG} occur more often than average, especially \texttt{DET}$\rightarrow$\texttt{VERB} (error rate of 10.0\% compared to 2.4(4.1)\% 
for other languages) and \texttt{DET}$\rightarrow$\texttt{PRON} (error rate of 13.5\% compared to 7.6(6.3)\% for other languages). 
\texttt{DET}$\rightarrow$\texttt{NOUN} is also common but similar to the other languages (error rate of 8.0\% compared to 7.0(7.0)\% for other languages). \texttt{DET}$\rightarrow$\texttt{VERB} and \texttt{DET}$\rightarrow$\texttt{NOUN} are negatively correlated for Wolof \carlos{ with an average coefficient of $-0.85(0.11)$ across all systems and all with p$<0.05$.}
Clearly, further language-specific analyses are needed. 
\section{Conclusion}
We have evaluated the impact POS tag accuracy has on parsing performance for leading graph- and transition-based parsers across a diverse range of UD treebanks, highlighting the stark difference between using predicted POS tags 
and gold POS tags at runtime. We observed a non-linear increase in performance when using gold tags, suggesting they are somehow exceptional\carlos{, i.e., precisely the tag patterns that not even the most accurate taggers can correctly predict (the last 2-3 percentage points towards 100\% accuracy) seem to be the most important for parsing}. \carlostwo{This could be due to the parsers implicitly learning POS tag information, in such a way that the taggers learn nothing new to contribute or not enough to avoid a loss in performance due to the errors disrupting what the parsers have learnt.}
Our analysis also shows that practitioners should evaluate the efficacy of using predicted tags for a given system or language. 
We have also analysed what aspects of erroneous tagging predictions have the greatest impact and correlation to parsing performance. We observed some global trends, \carlos{like the importance of \texttt{CCONJ},} but also language-specific issues which highlight the need to evaluate the usefulness of POS tags per language. The results also suggest that using a subset of POS tags might be effective.

\section*{Acknowledgments}

\carlostwo{This work has received funding from the European Research Council (ERC), under the European Union's Horizon 2020 research and innovation programme (FASTPARSE, grant agreement No 714150), from MINECO (ANSWER-ASAP, TIN2017-85160-C2-1-R), from Xunta de Galicia (ED431C 2020/11), and from Centro de Investigación de Galicia ``CITIC'', funded by Xunta de Galicia and the European Union (ERDF - Galicia 2014-2020 Program), by grant ED431G 2019/01. The authors would also like to the thank \textit{all} the reviewers for their detailed and constructive comments and criticisms.}

\bibliography{emnlp-ijcnlp-2019}
\bibliographystyle{acl_natbib}
\appendices
\clearpage
\onecolumn
\section{Treebank performances}\label{appendix:tbs}

\begin{figure}[hb!]
    \centering
    \includegraphics[width=0.98\linewidth]{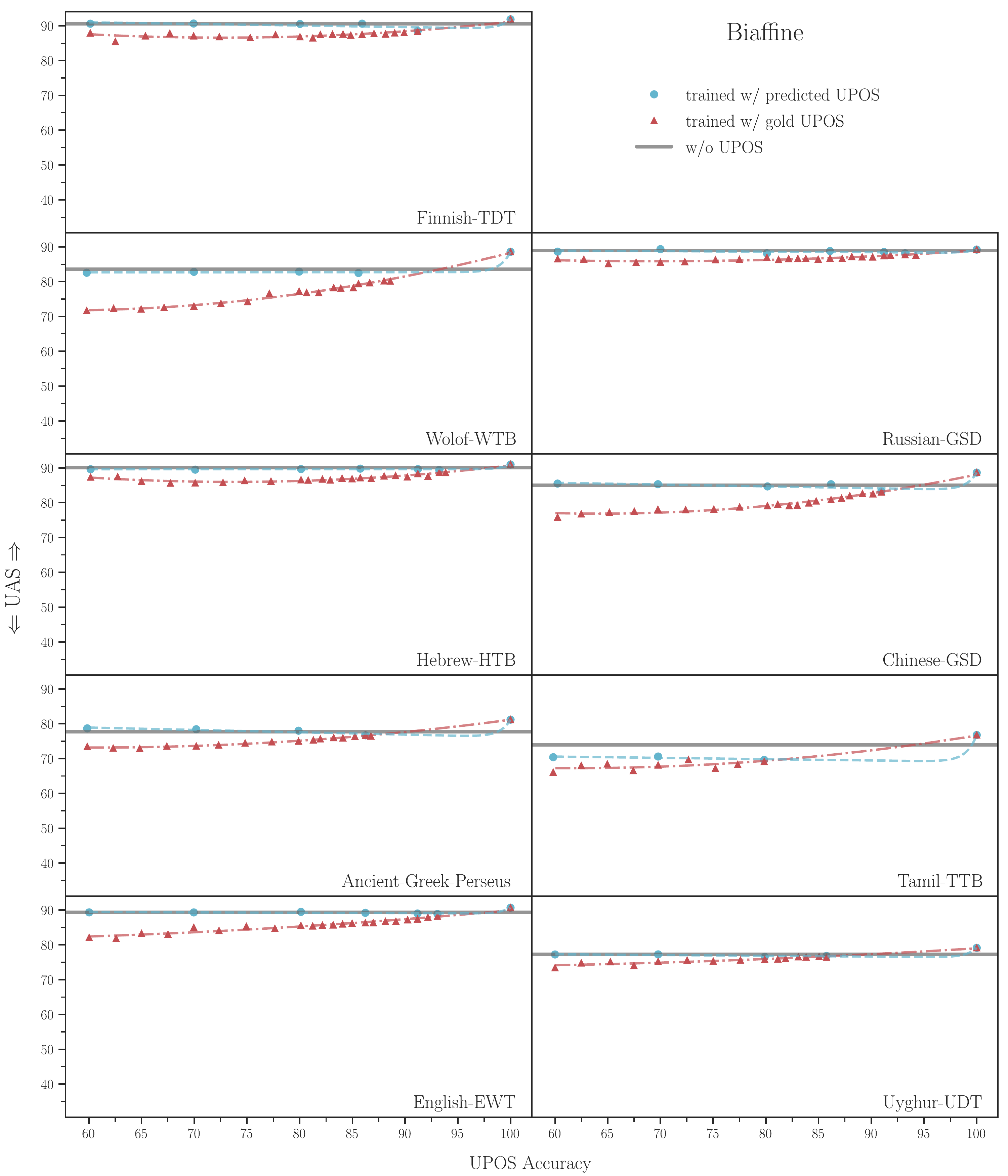}
    \caption{UAS for each treebank for Biaffine training with gold (red, triangles) and predicted (blue, circles) POS tags. Baseline parser trained without POS tags is shown in grey.}
    \label{fig:uas_biaffine}
\end{figure}
\clearpage
\begin{figure}[h!]
    \centering
    \includegraphics[width=0.98\linewidth]{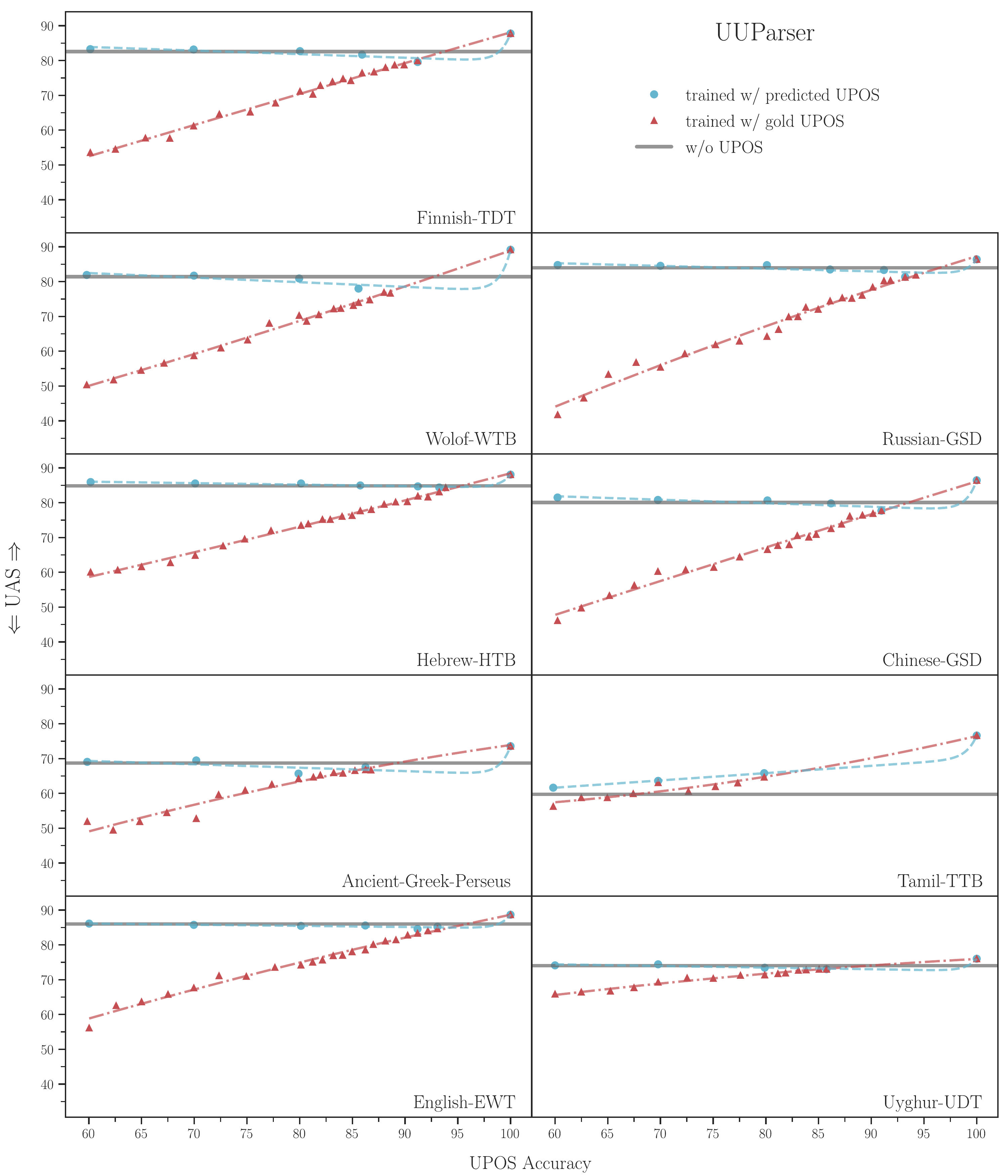}
    \caption{UAS for each treebank for UUParser training with gold (red, triangles) and predicted (blue, circles) POS tags. Baseline parser trained without POS tags is shown in grey.}
    \label{fig:uas_uppsala}
\end{figure}
\clearpage
\begin{figure}[h!]
    \centering
    \includegraphics[width=0.98\linewidth]{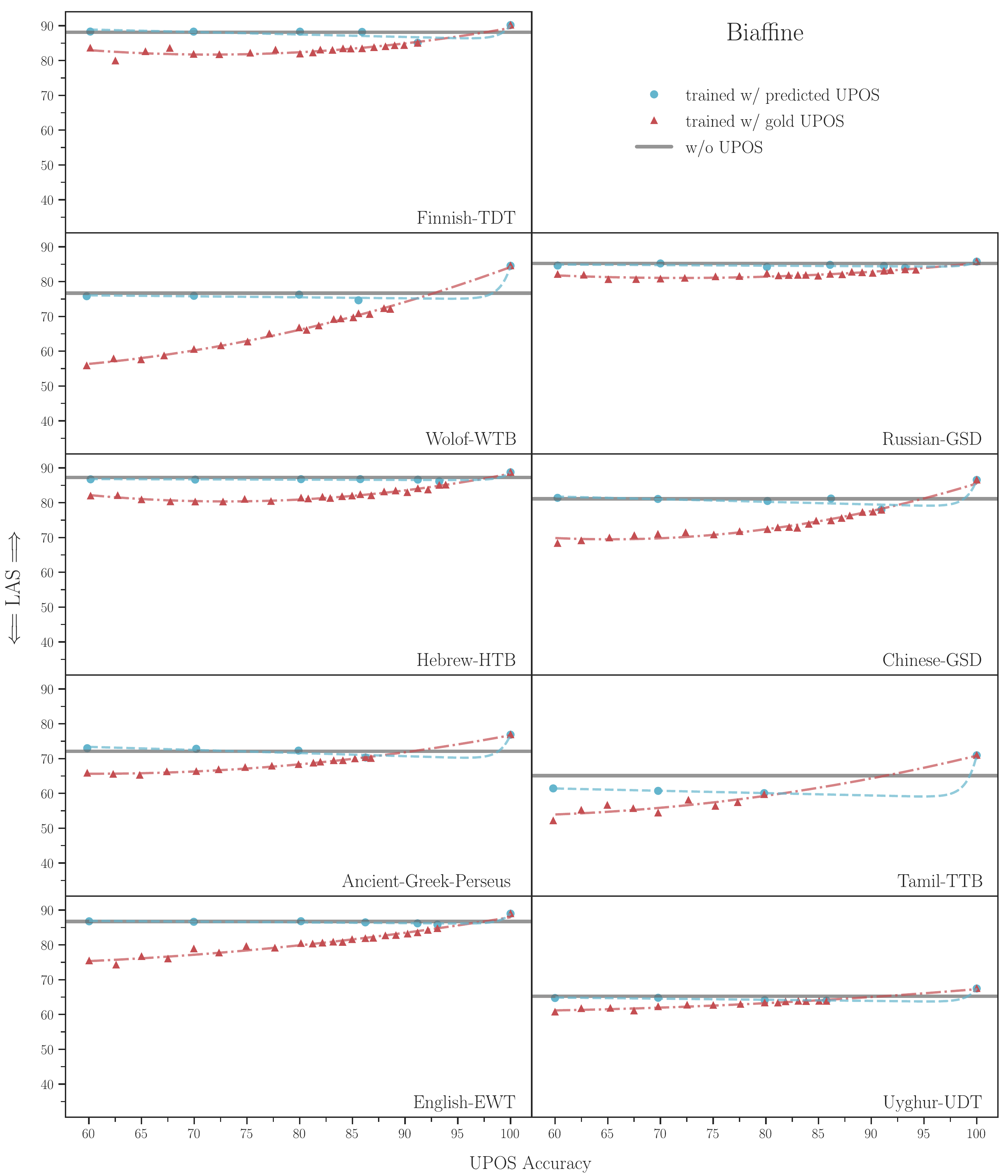}
    \caption{LAS for each treebank for Biaffine training with gold (red, triangles) and predicted (blue, circles) POS tags. Baseline parser trained without POS tags is shown in grey.}
    \label{fig:las_biaffine}
\end{figure}
\clearpage
\begin{figure}[h!]
    \centering
    \includegraphics[width=0.98\linewidth]{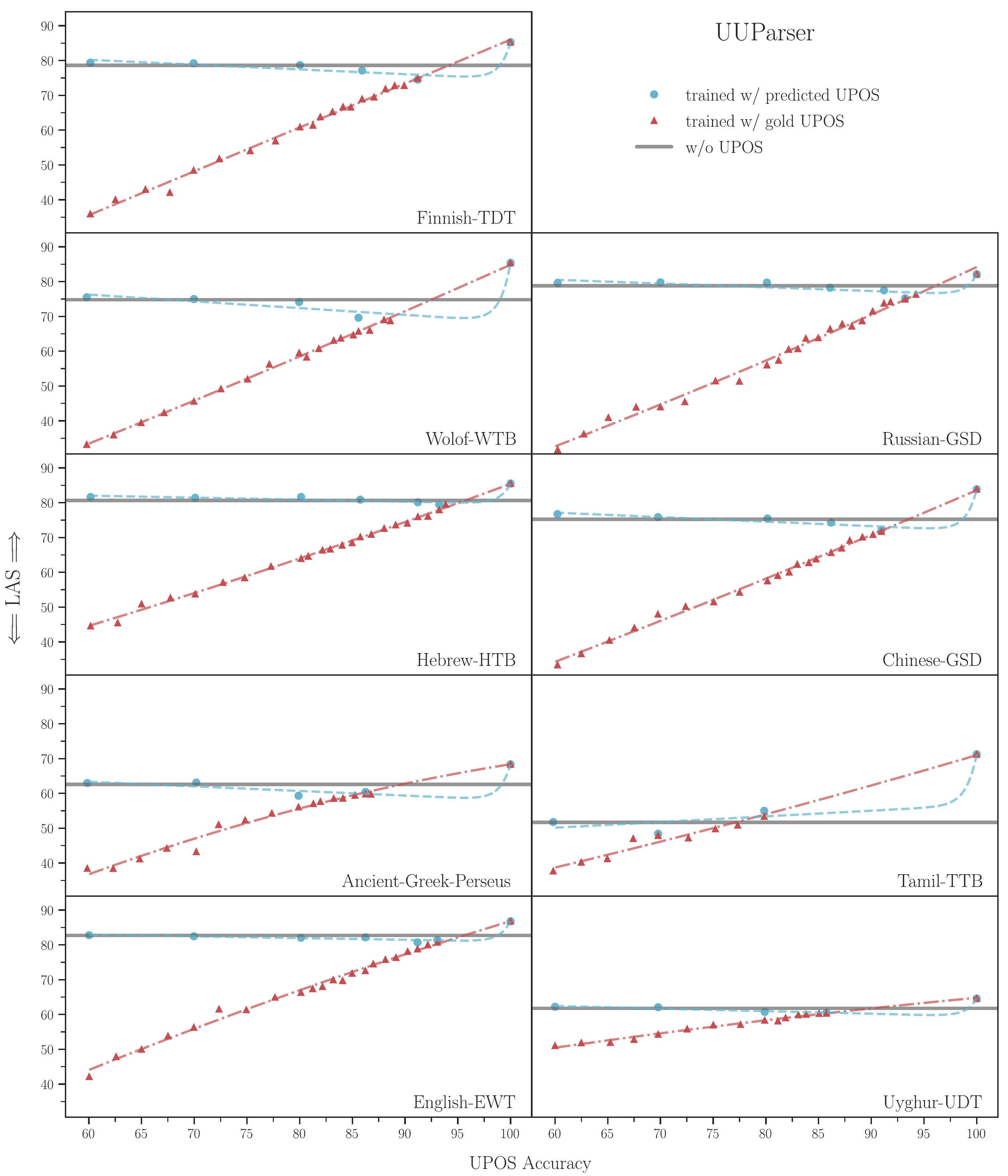}
    \caption{LAS for each treebank for UUParser training with gold (red, triangles) and predicted (blue, circles) POS tags. Baseline parser trained without POS tags is shown in grey.}
    \label{fig:las_uppsala}
\end{figure}

\begin{figure}[tb]
    \centering
    \includegraphics[width=0.98\linewidth]{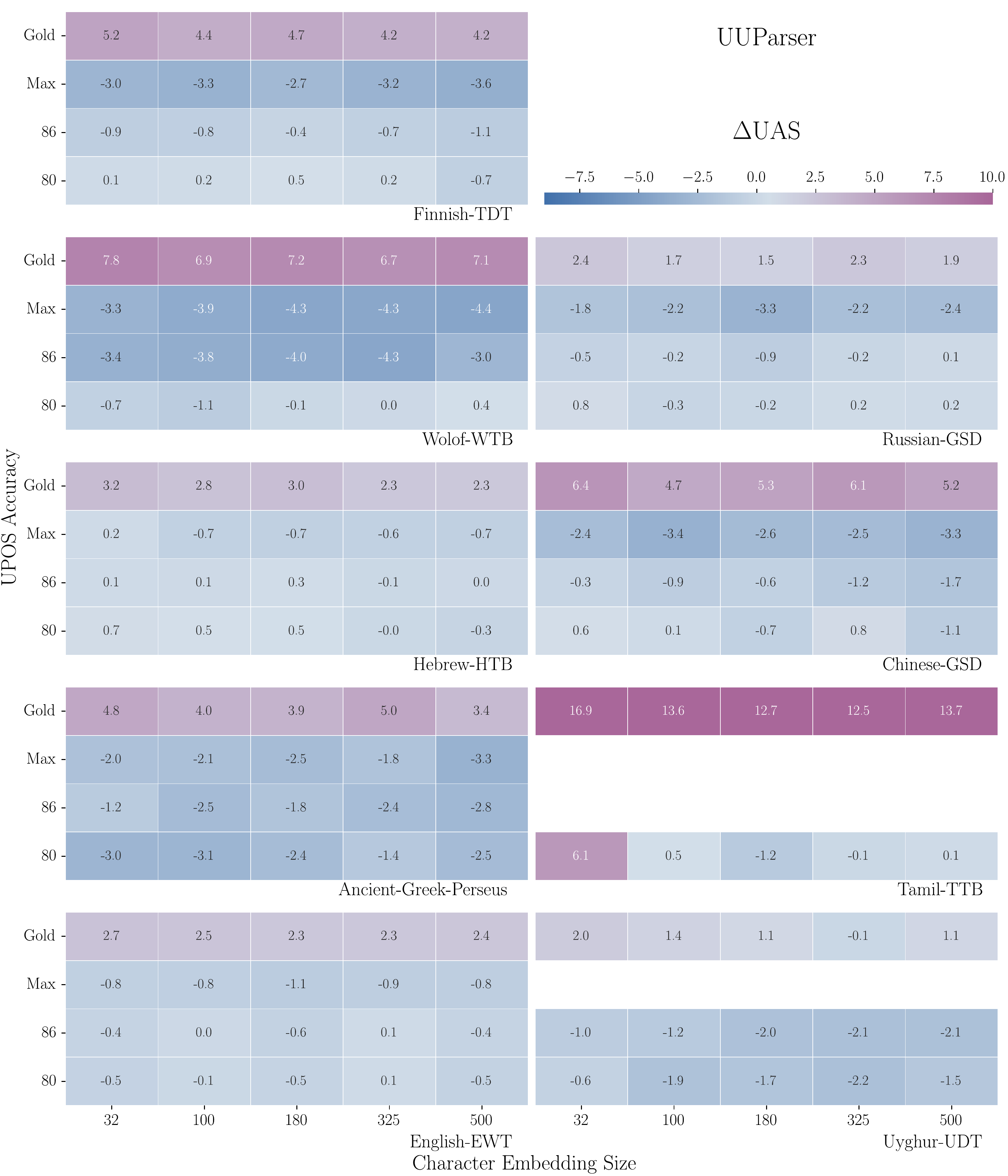}
    \caption{$\Delta$ UAS for each treebank for UUParser compared to the baseline parsers trained without POS tags for different character embedding sizes and different POS tag accuracies.}
    \label{fig:uas_uppsala_heat}
\end{figure}
\begin{figure}[tb]
    \centering
    \includegraphics[width=0.98\linewidth]{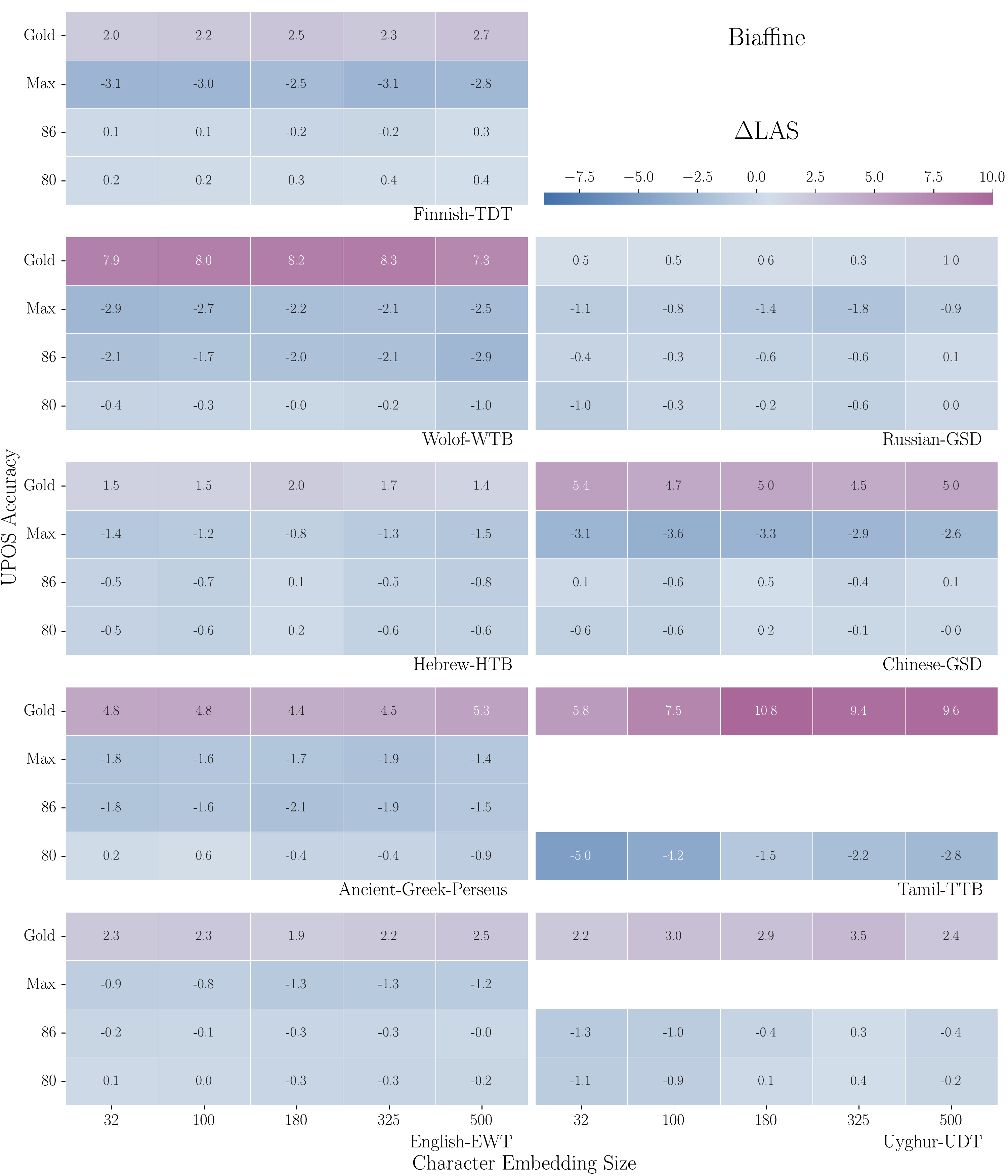}
    \caption{$\Delta$ LAS for each treebank for Biaffine compared to the baseline parsers trained without POS tags for different character embedding sizes and different POS tag accuracies.}
    \label{fig:las_biaffine_heat}
\end{figure}

\begin{figure}[tb]
    \centering
    \includegraphics[width=0.98\linewidth]{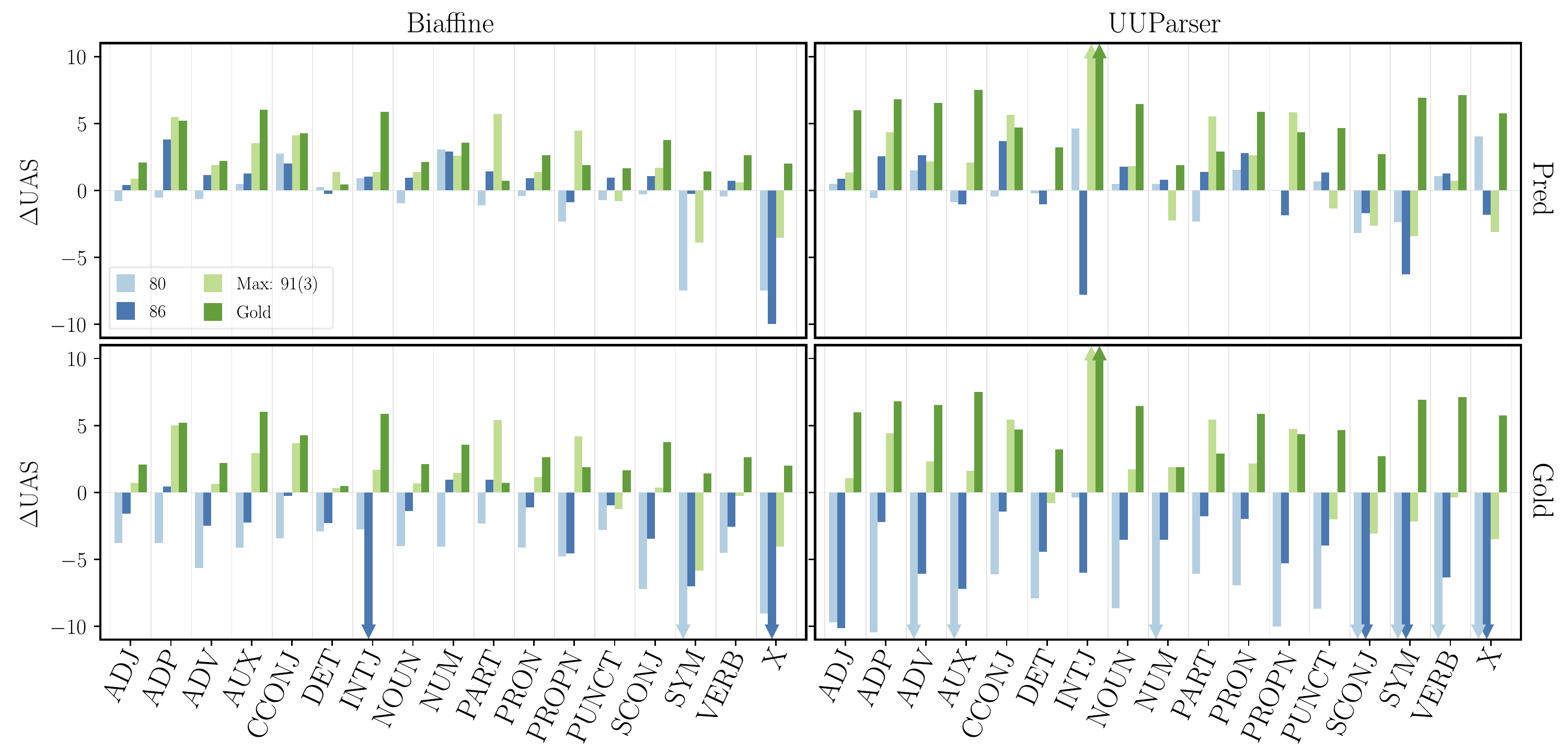}
    \caption{Average $\Delta$UAS across all treebanks for models trained with POS tags from taggers of 80, 86, max POS accuracy of 91(3), and with gold tags for each POS tag.}
    \label{fig:duas_bar}
\end{figure}
\begin{figure}[tb]
    \centering
    \includegraphics[width=0.98\linewidth]{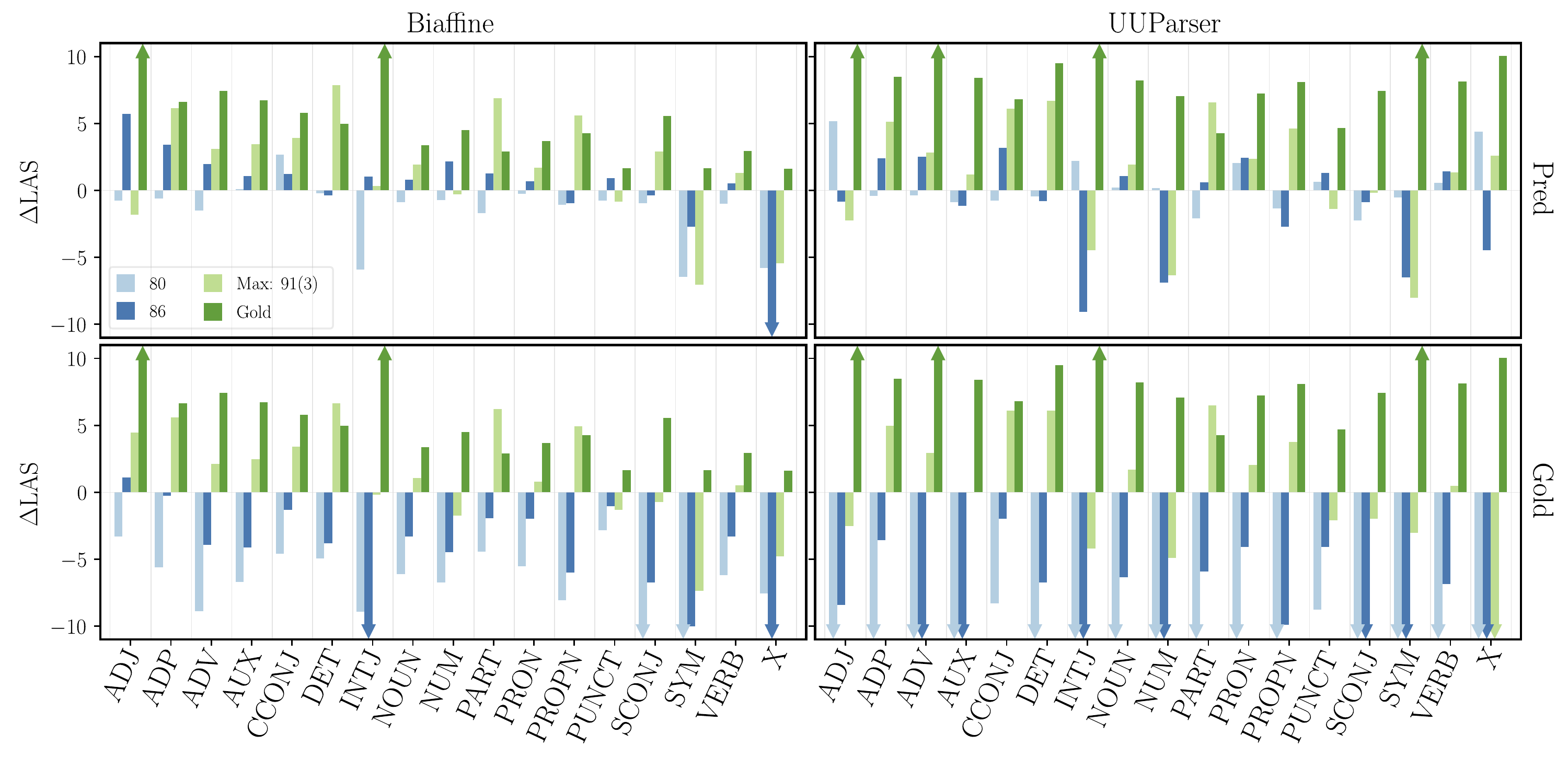}
    \caption{Average $\Delta$LAS across all treebanks for models trained with POS tags from taggers of 80, 86, max POS accuracy of 91(3), and with gold tags for each POS tag.}
    \label{fig:dlas_bar}
\end{figure}

\begin{figure}[tb]
    \centering
    \includegraphics[width=0.98\linewidth]{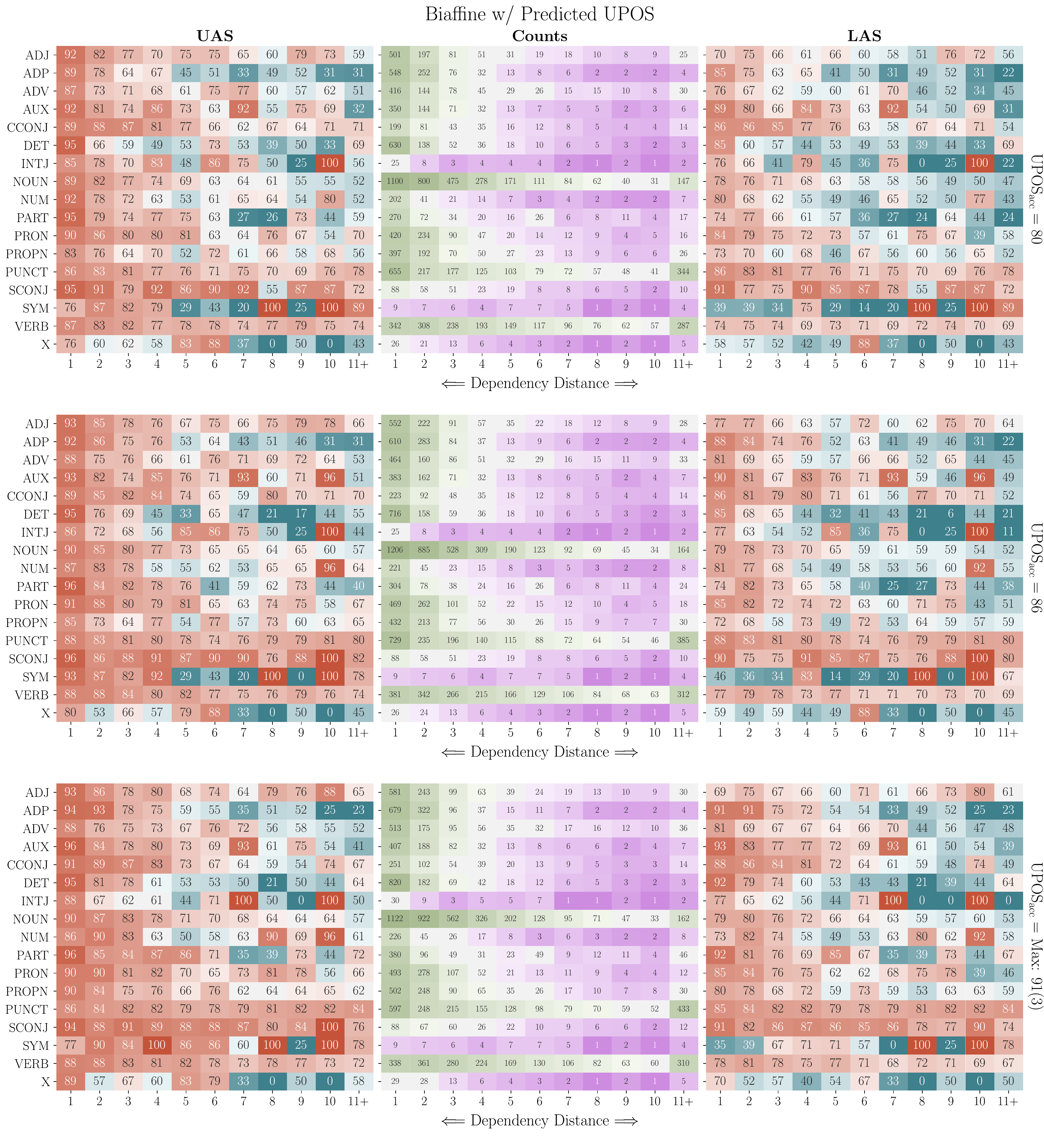}
    \caption{Average UAS (left column) and LAS (right column) across treebanks with Biaffine for models trained with predicted tags from taggers with 80, 86, and max POS accuracy of 91(3). UAS and LAS metrics are shown for each gold tag (y-axis) with a given dependency distance (x-axis). The counts of each pair of POS tag and dependency distance are shown in the middle column.}
    \label{fig:uas_dep_dist_biaffine}
\end{figure}

\begin{figure}[tb]
    \centering
    \includegraphics[width=0.98\linewidth]{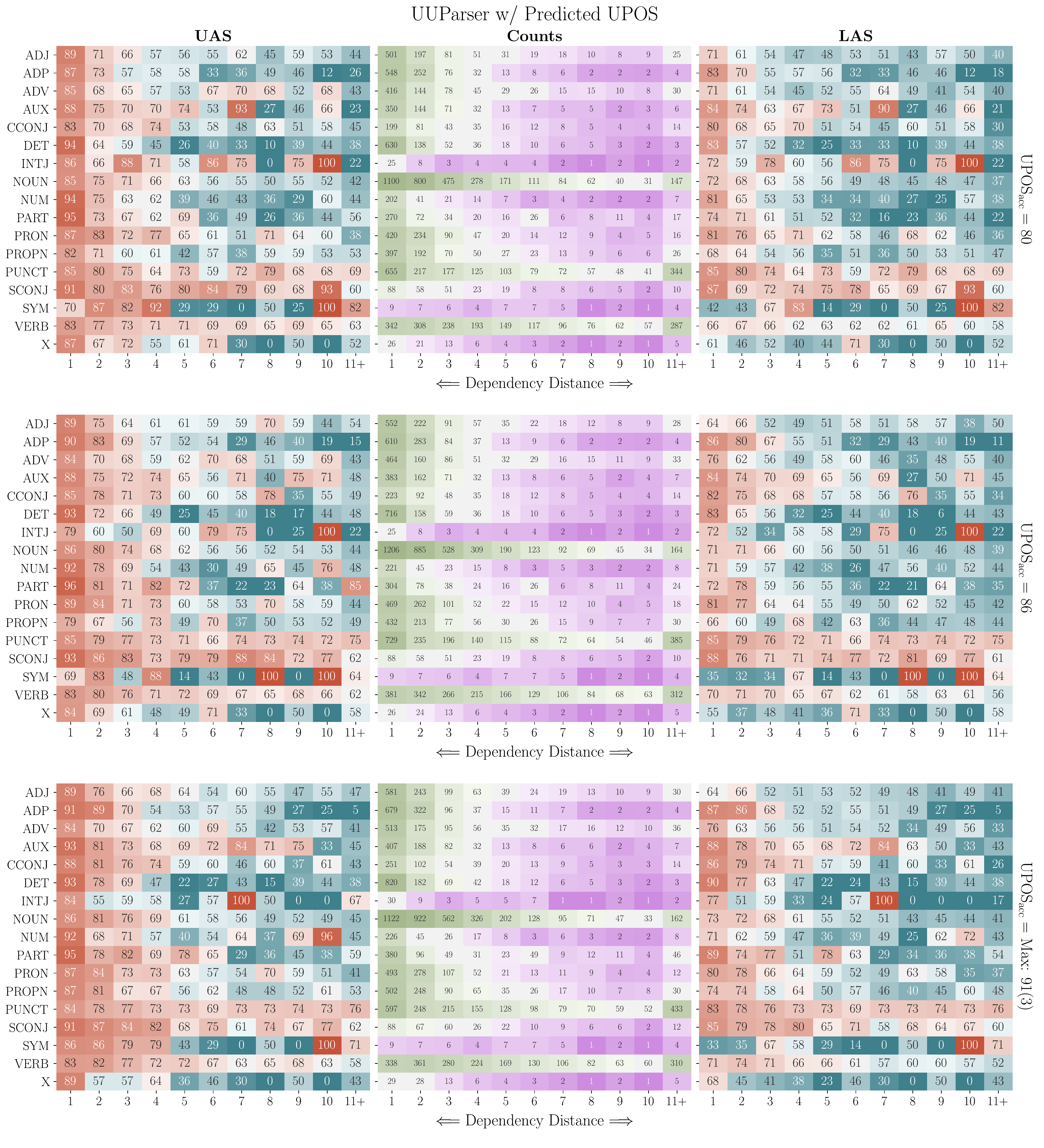}
    \caption{Average UAS (left column) and LAS (right column) across treebanks with UUParser for models trained with predicted tags from taggers with 80, 86, and max POS accuracy of 91(3). UAS and LAS metrics are shown for each gold tag (y-axis) with a given dependency distance (x-axis). The counts of each pair of POS tag and dependency distance are shown in the middle column.}
    \label{fig:uas_dep_dist_uppsala}
\end{figure}

\begin{figure}[tb]
    \centering
    \includegraphics[width=0.98\linewidth]{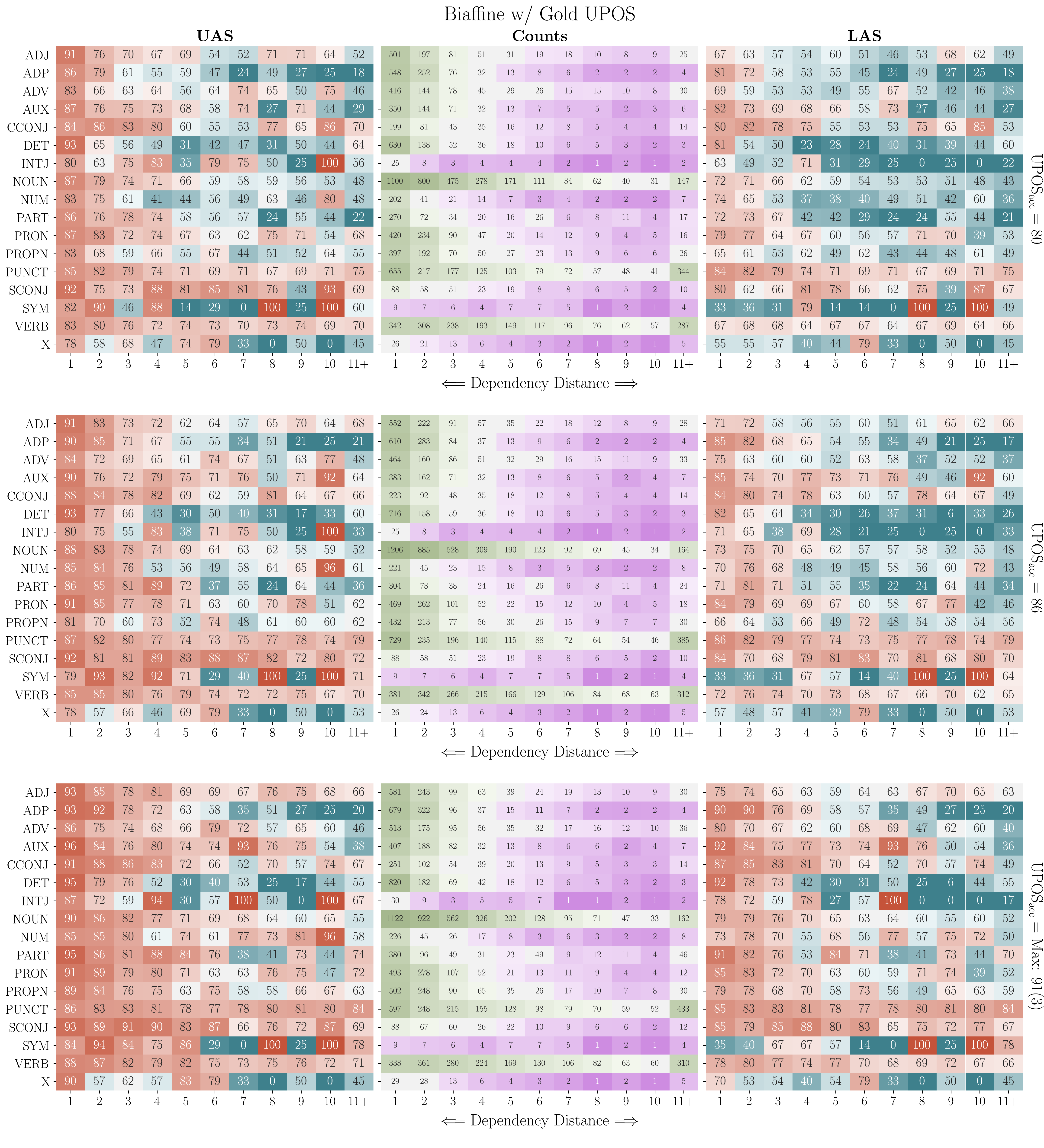}
    \caption{Average UAS (left column) and LAS (right column) across treebanks with Biaffine for models trained with gold tags but using predicted tags from taggers with 80, 86, and max POS accuracy of 91(3). UAS and LAS metrics are shown for each gold tag (y-axis) with a given dependency distance (x-axis). The counts of each pair of POS tag and dependency distance are shown in the middle column.}
    \label{fig:uas_dep_dist_biaffine_gold}
\end{figure}

\begin{figure}[tb]
    \centering
    \includegraphics[width=0.98\linewidth]{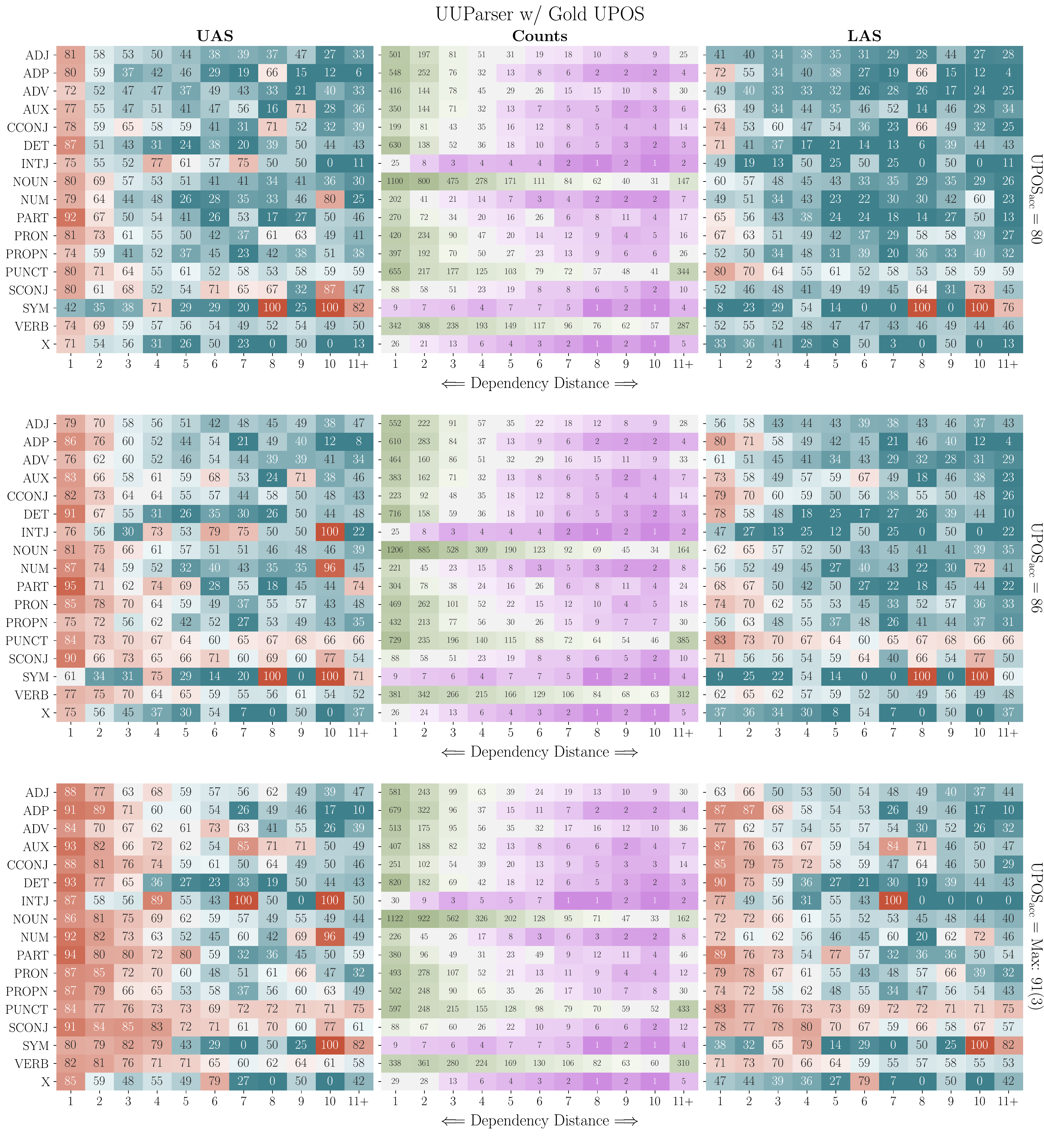}
    \caption{Average UAS (left column) and LAS (right column) across treebanks with UUParser for models trained with gold tags but using predicted tags from taggers with 80, 86, and max POS accuracy of 91(3). UAS and LAS metrics are shown for each gold tag (y-axis) with a given dependency distance (x-axis). The counts of each pair of POS tag and dependency distance are shown in the middle column.}
    \label{fig:uas_dep_dist_uppsala_gold}
\end{figure}
\begin{figure}[tb]
    \centering
    \includegraphics[width=0.98\linewidth]{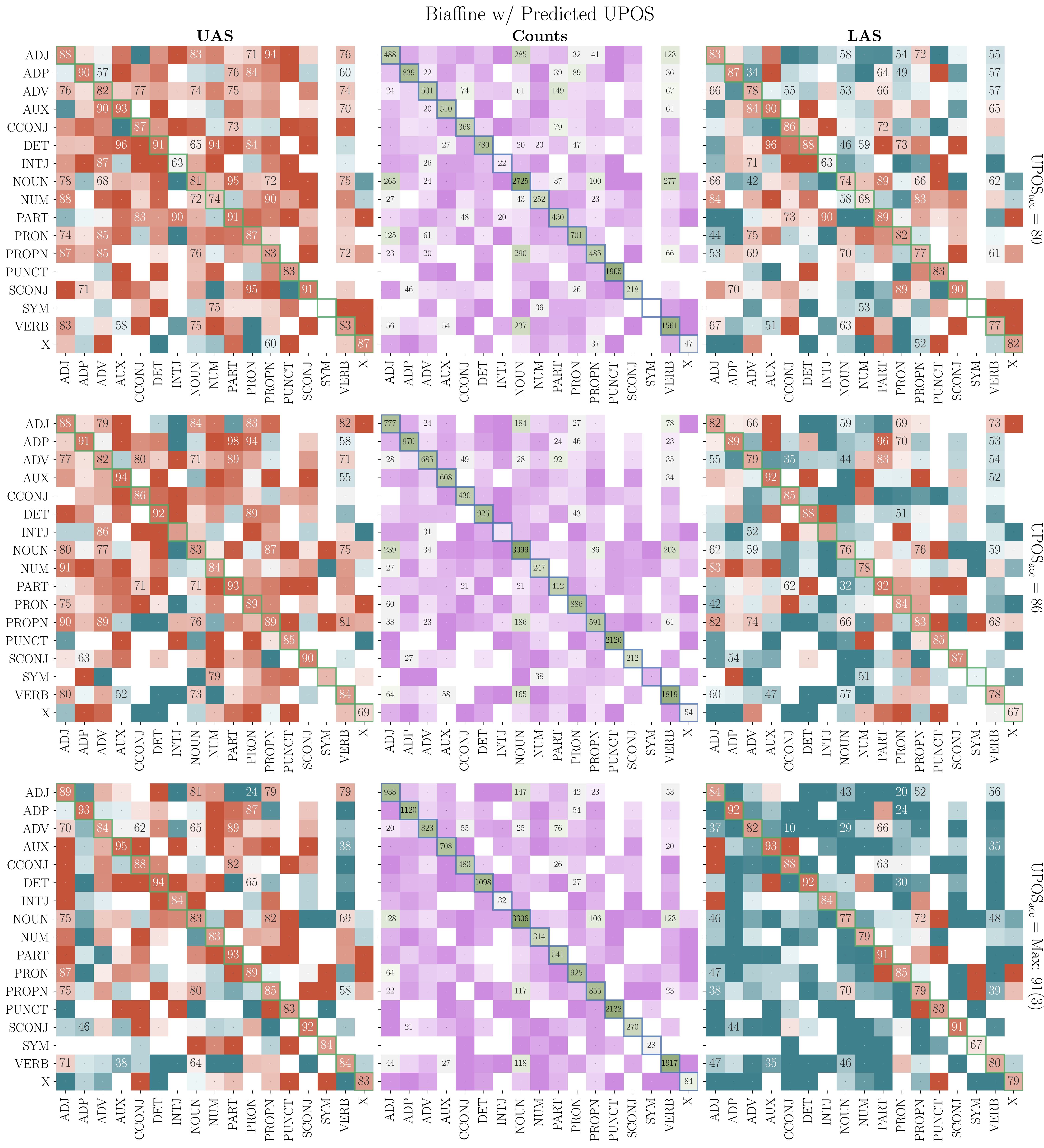}
    \caption{Average UAS (left column) and LAS (right column) across treebanks with Biaffine for models trained with POS tags from taggers of 80, 86, and max POS accuracy of 91(3). UAS and LAS metrics are shown for each gold tag (y-axis) predicted as any other tag (x-axis). The numbers are annotated when the average count (shown in the centre column) of a particular error is greater than 20.}
    \label{fig:uas_x_as_y_biaffine}
\end{figure}

\begin{figure}[tb]
    \centering
    \includegraphics[width=0.98\linewidth]{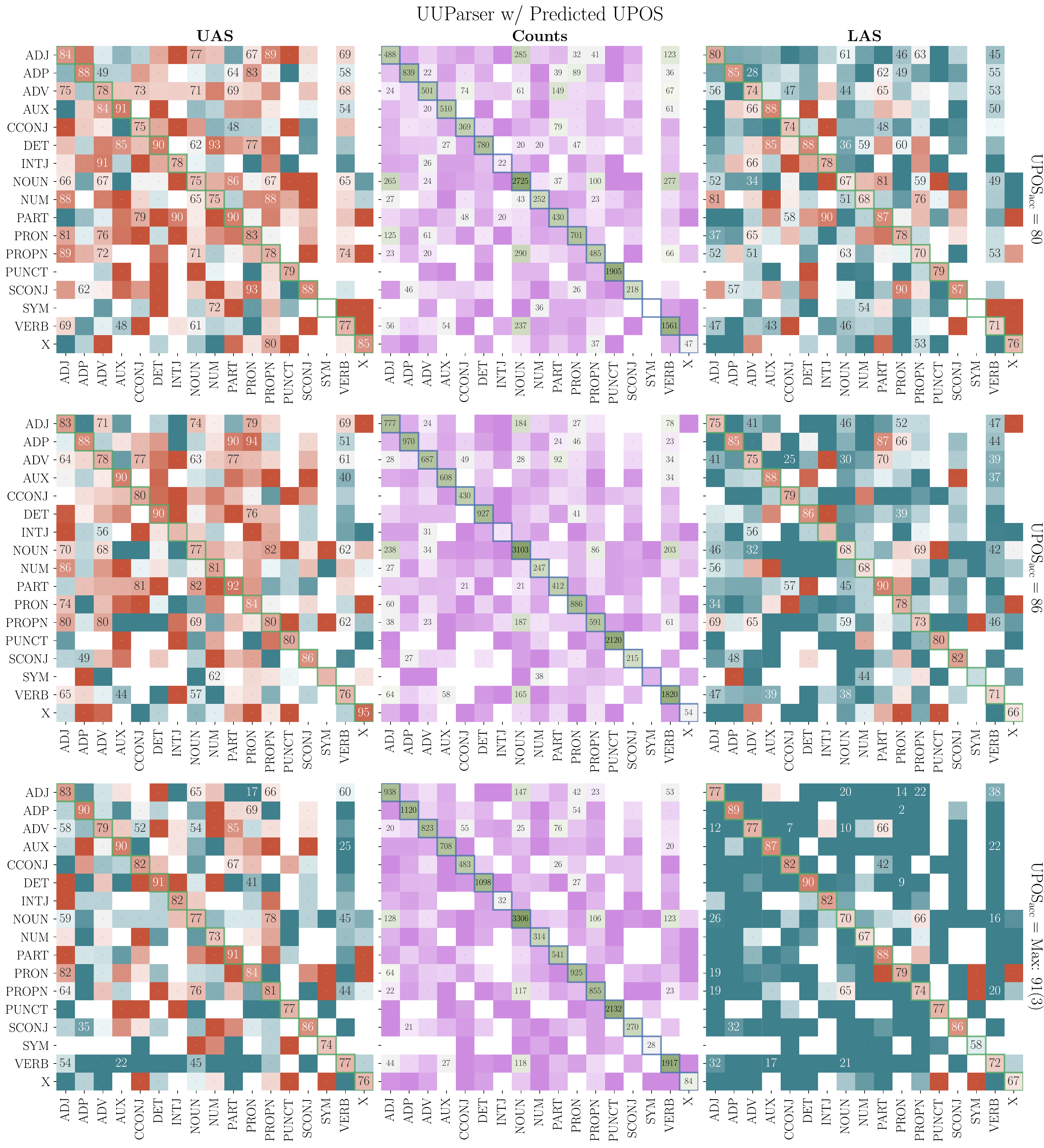}
    \caption{Average UAS (left column) and LAS (right column) across treebanks with UUParser for models trained with POS tags from taggers of 80, 86, and max POS accuracy of 91(3). UAS and LAS metrics are shown for each gold tag (y-axis) predicted as any other tag (x-axis). The numbers are annotated when the average count (shown in the centre column) of a particular error is greater than 20.}
    \label{fig:uas_x_as_y_uppsala}
\end{figure}

\begin{figure}[tb]
    \centering
    \includegraphics[width=0.98\linewidth]{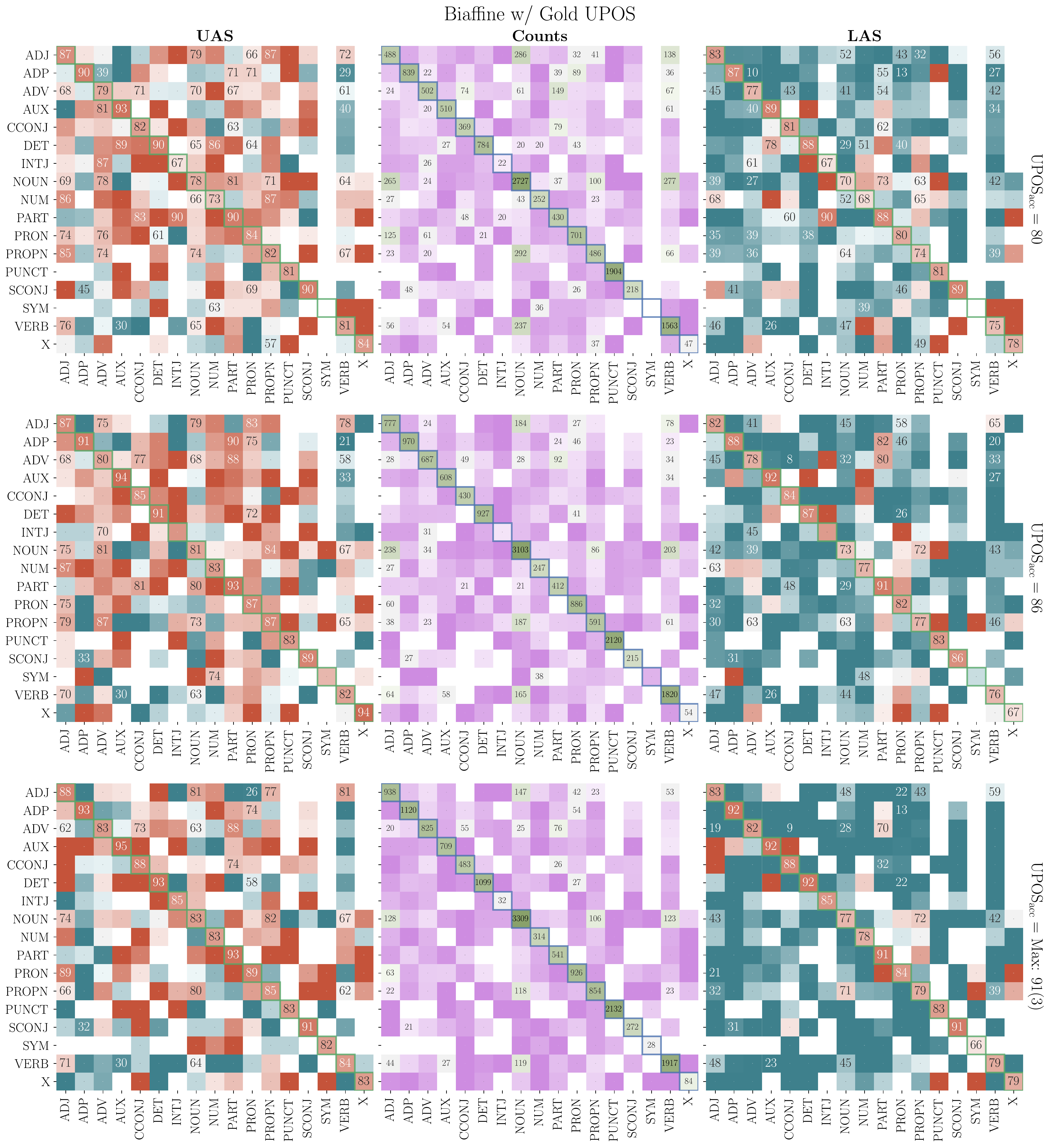}
    \caption{Average UAS (left column) and LAS (right column) across treebanks with Biaffine for models trained with gold tags but using predicted tags from taggers with 80, 86, and max POS accuracy of 91(3). UAS and LAS metrics are shown for each gold tag (y-axis) predicted as any other tag (x-axis). The numbers are annotated when the average count (shown in the centre column) of a particular error is greater than 20.}
    \label{fig:uas_x_as_y_biaffine_gold}
\end{figure}

\begin{figure}[tb]
    \centering
    \includegraphics[width=0.98\linewidth]{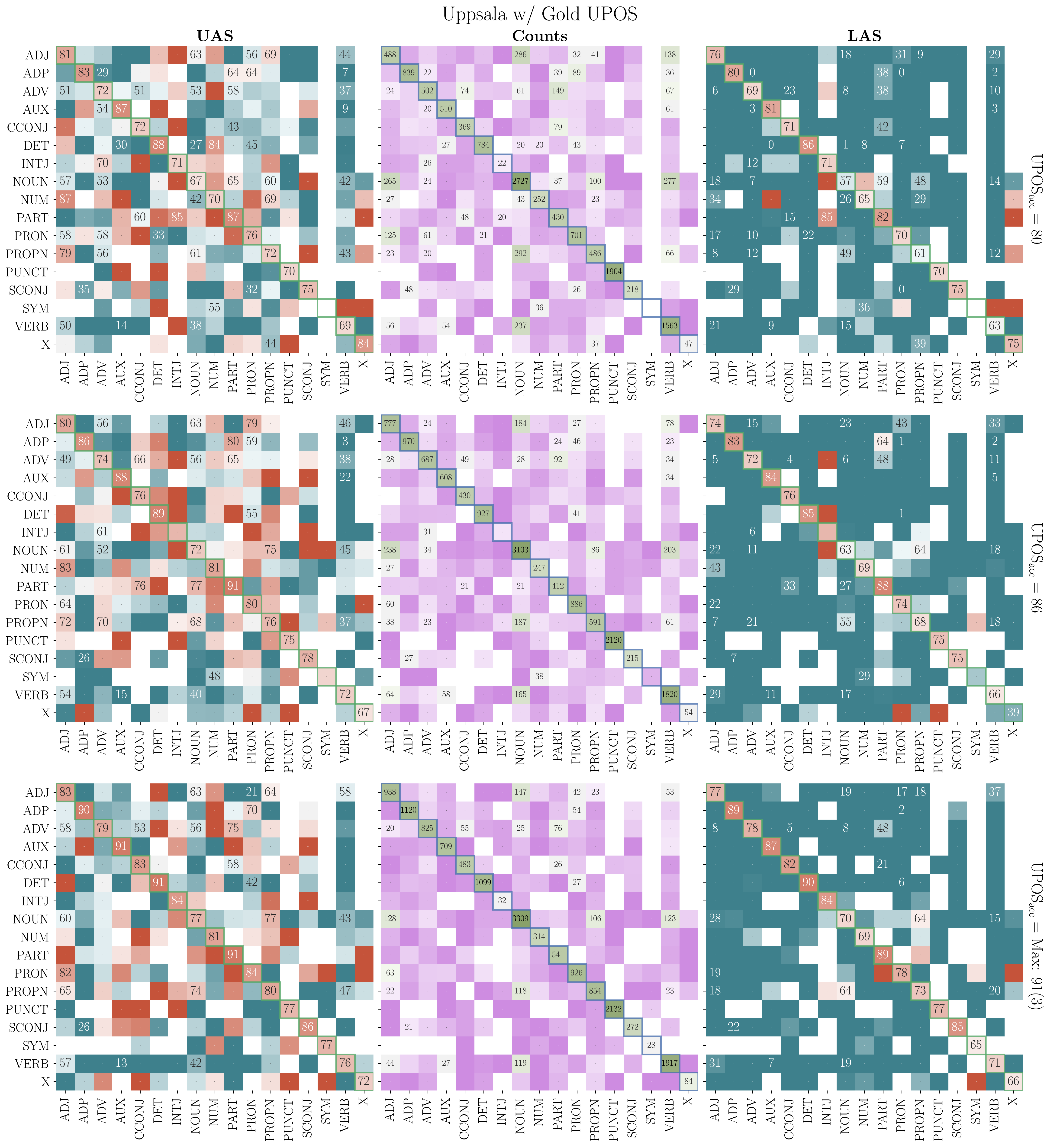}
    \caption{Average UAS (left column) and LAS (right column) across treebanks with UUParser for models trained with gold tags but using predicted tags from taggers with 80, 86, and max POS accuracy of 91(3). UAS and LAS metrics are shown for each gold tag (y-axis) predicted as any other tag (x-axis). The numbers are annotated when the average count (shown in the centre column) of a particular error is greater than 20.}
    \label{fig:uas_x_as_y_uppsala_gold}
\end{figure}


\clearpage
\begin{figure}[htbp!]
    \centering
    \includegraphics[width=0.98\linewidth]{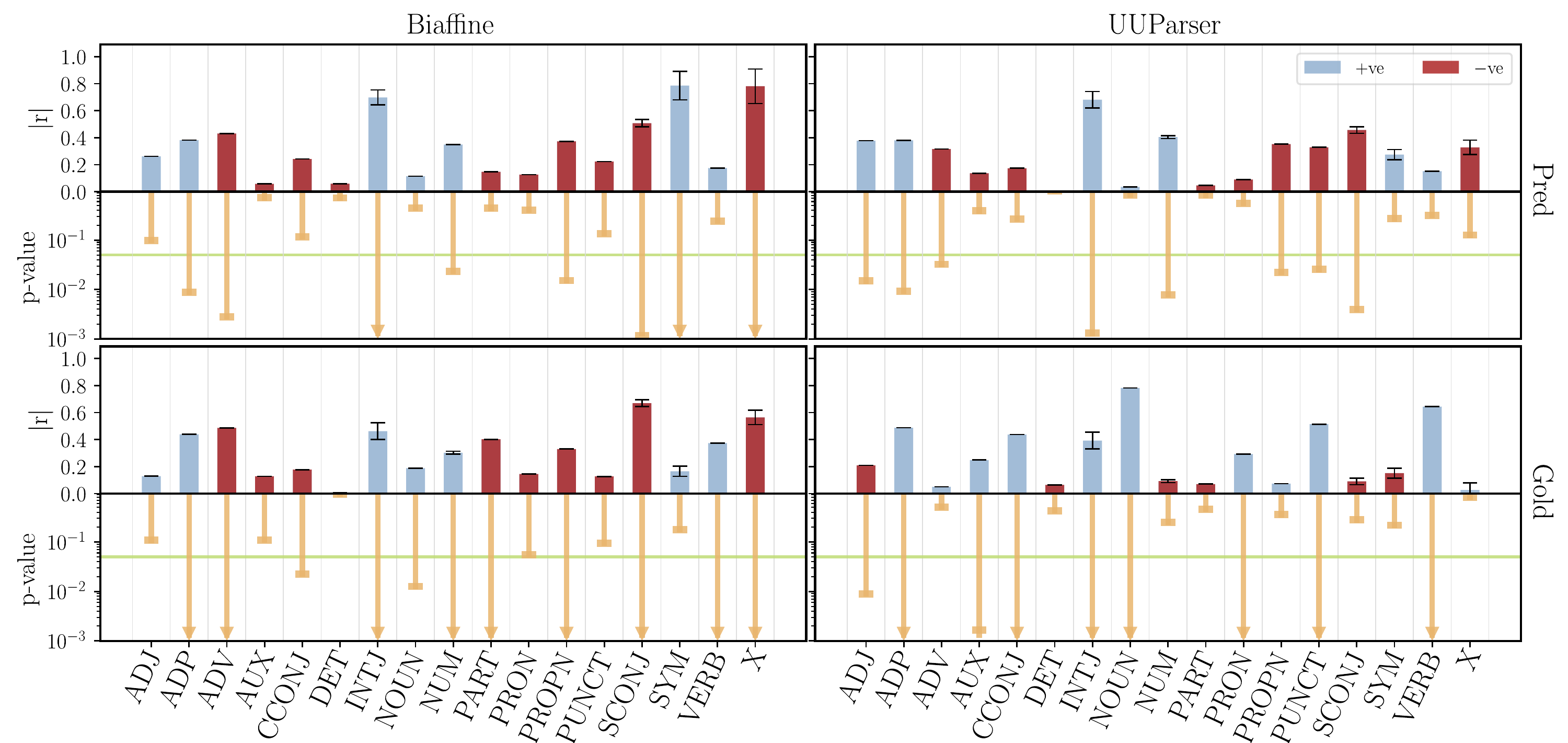}
    \caption{Pearson coefficients for the F1-score of the \textit{head} of separate POS tags and global LAS where positive (+ve) coefficients are shown in blue and negative (-ve) are shown in red. The corresponding p-values are shown below (orange) where an arrow head means the value was below 0.001. Left subplots are for Biaffine parsers, right for UUParsers, top row is for parsers trained with predicted tags, and bottom for parsers trained with gold tags.}
    \label{fig:corr_head}
\end{figure}
\begin{figure}[htpb!]
    \centering
    \includegraphics[width=0.98\linewidth]{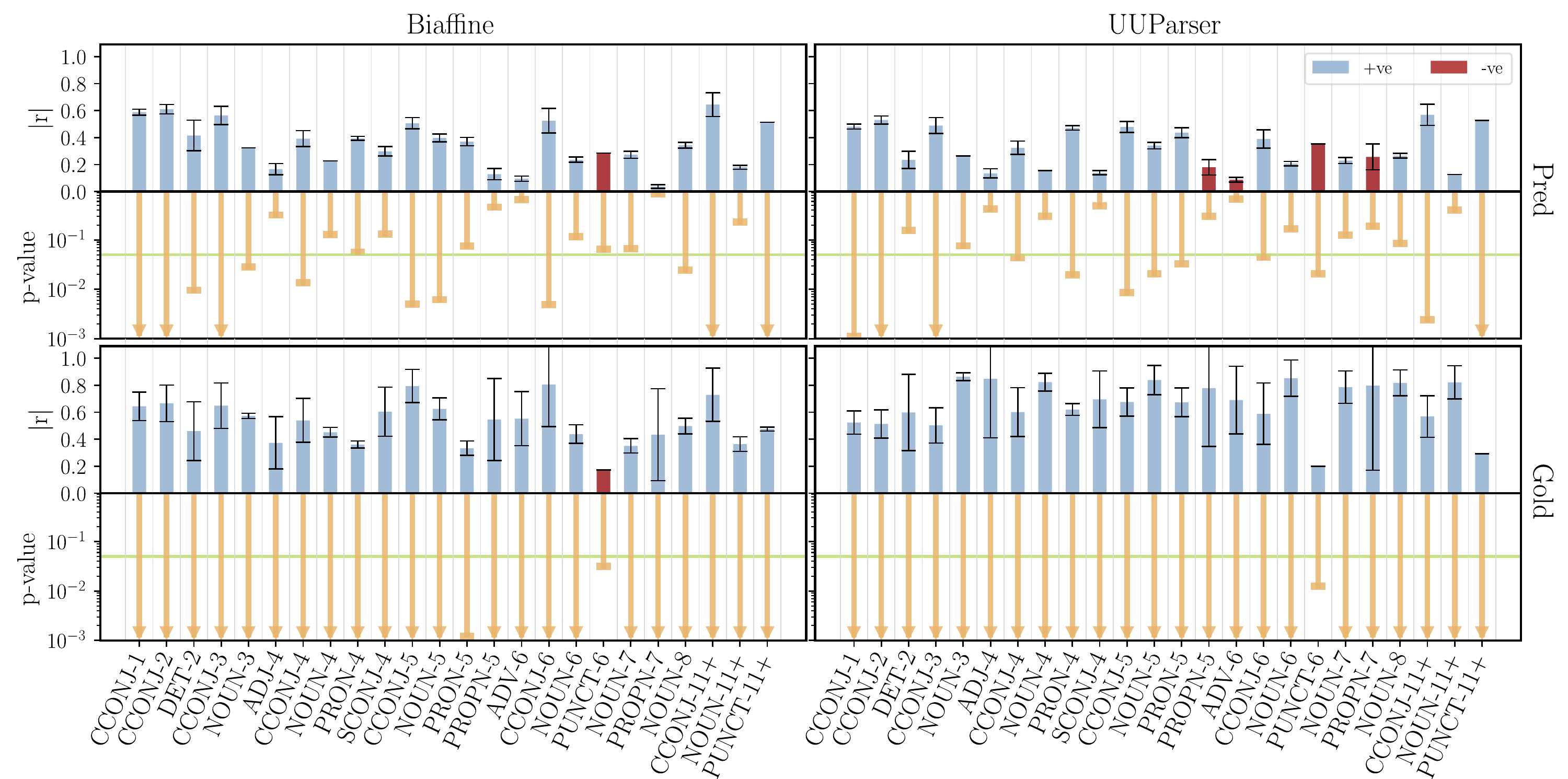}
    \caption{Pearson coefficients for the F1-score for individual POS tags with different dependency distances (\texttt{POS}-distance) and global LAS where positive (+ve) coefficients are shown in blue and negative (-ve) are shown in red. The corresponding p-values are shown below (orange) where an arrow head means the value was below 0.001. Left subplots are for Biaffine parsers, right for UUParsers, top row is for parsers trained with predicted tags, and bottom for parsers trained with gold tags.}
    \label{fig:corr_dist}
\end{figure}
\end{document}